\pdfoutput=1

\documentclass[11pt]{article}

\PassOptionsToPackage{dvipsnames}{xcolor}
\usepackage{acl}
\usepackage{marvosym}
\usepackage{times}
\usepackage{latexsym}
\usepackage{graphicx}
\usepackage{bm}
\usepackage[T1]{fontenc}

\usepackage[utf8]{inputenc}

\usepackage{microtype}
\usepackage{float}
\usepackage{amsmath, amssymb}
\usepackage{adjustbox}
\usepackage{stfloats}

\usepackage[dvipsnames]{xcolor} 
\usepackage{pifont}            

\newcommand{\cmark}{\textcolor{ForestGreen}{\ding{51}}} 
\newcommand{\xmark}{\textcolor{Red}{\ding{55}}}        

\usepackage{booktabs} 
\usepackage{multirow} 

\usepackage{enumitem}

\def\eqref#1{equation~\ref{#1}}

\def\1{\bm{1}}

\def\vh{{\bm{h}}}

\def\vu{{\bm{u}}}
\def\vv{{\bm{v}}}

\def\vz{{\bm{z}}}

\def\mW{{\bm{W}}}

\DeclareMathAlphabet{\mathsfit}{\encodingdefault}{\sfdefault}{m}{sl}
\SetMathAlphabet{\mathsfit}{bold}{\encodingdefault}{\sfdefault}{bx}{n}

\def\gD{{\mathcal{D}}}

\def\gH{{\mathcal{H}}}

\def\gP{{\mathcal{P}}}

\def\gS{{\mathcal{S}}}

\usepackage{xspace}
\newcommand{\method}{SITA\xspace}

\usepackage{hyperref}
\usepackage{cleveref}

\newcommand{\draftonly}[1]{}

\title{\method: Learning Speaker-Invariant and Tone-Aware Speech Representations for Low-Resource Tonal Languages}

\author{
{\large\bfseries
Tianyi Xu\textsuperscript{1,*}\quad
Xuan Ouyang\textsuperscript{1,*}\quad
Binwei Yao\textsuperscript{1}\quad
Shoua Xiong\textsuperscript{2}}\\[0.1em]
{\large\bfseries
Sara Misurelli\textsuperscript{3}\quad
Maichou Lor\textsuperscript{2}\quad
Junjie Hu\textsuperscript{1,\protect\Letter}}\\[0.35em]
{\large
\textsuperscript{1}Department of Computer Sciences\quad
\textsuperscript{2}School of Nursing\quad
\textsuperscript{3}Department of Otolaryngology}\\
{\large University of Wisconsin--Madison, Madison, WI, USA}\\[0.25em]
{\large \textsuperscript{*}Equal contribution\qquad
\textsuperscript{\protect\Letter}Corresponding author}
}

\begin{document}
\maketitle

\begin{abstract}
Tonal low-resource languages are widely spoken but remain underserved by modern speech technologies. A central challenge is learning speech representations that are robust to nuisance variation, such as speaker gender, while preserving lexical tone, which carries word meaning. We propose \textbf{\method}, a lightweight adaptation recipe for pretrained wav2vec-style self-supervised speech encoders. Rather than designing a new backbone or objective, \method combines existing objectives in a staged optimization framework to reduce tone collapse while preserving ASR capability.
Stage 1 improves speaker invariance without erasing tonal contrasts by combining a cross-gender contrastive loss with a tone-repulsive loss that separates same-word, different-tone realizations. Stage 2 restores recognition-oriented linguistic information through CTC fine-tuning and knowledge distillation on upper encoder layers.
We evaluate \method primarily on Hmong, a tonal language with limited digital resources and a small speaker pool. Against multilingual, speaker-adversarial, label-aware, and semi-supervised baselines, \method achieves the best trade-off between cross-gender lexical retrieval and tone separation, while maintaining ASR accuracy close to an ASR-adapted XLS-R teacher. Results on Mandarin show consistent gains, suggesting that \method is a general plug-in recipe for tonal speech representation learning.
\end{abstract}

\section{Introduction}

\begin{figure}[t]
    \centering
    \includegraphics[width=\linewidth]{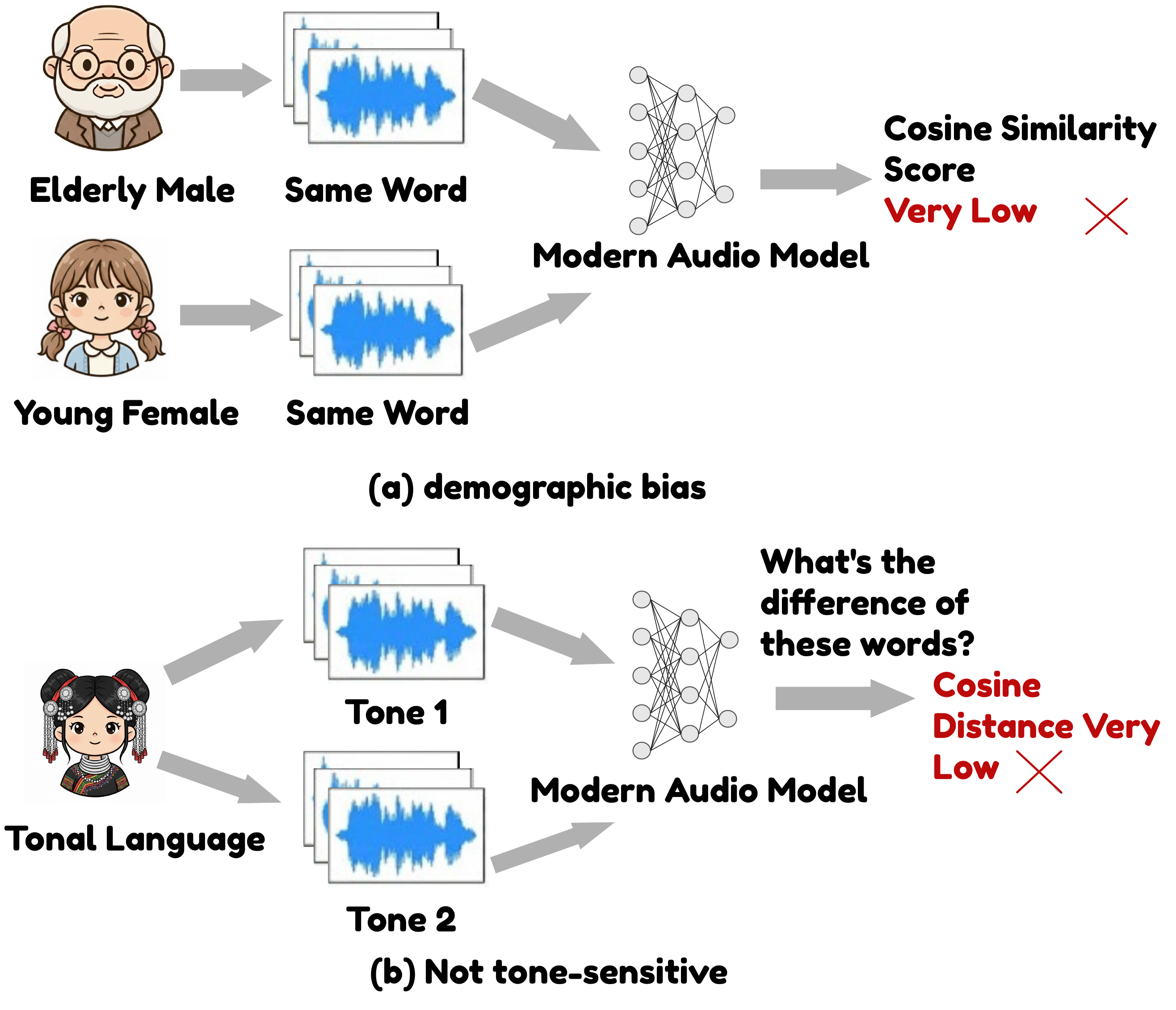}
    \vspace{-20pt}
    \caption{Two failure modes of speech embedding collapse: (1) demographic bias, where embeddings of the same word spoken by different speakers are insufficiently similar, and (2) weak tone sensitivity, where embeddings of the same base word carrying different tones are not meaningfully distinguishable. 
    }
    \vspace{-20pt}
    \label{fig:intro_failure}
\end{figure}
Tonal languages pose a core challenge for speech representation learning. Models must be sensitive to lexical tone, which encodes meaning, while remaining robust to nuisance variation such as speaker identity and recording conditions \cite{shen2024encoding, besacier2014automatic}. Here lexical tone refers to pitch contours that change word meaning rather than speaker accent, so two utterances with the same segmental content but different tones are different words. Although over half of the world’s languages are tonal, they remain underrepresented in speech resources, with scarce labels and strong speaker imbalance \cite{besacier2014automatic, singh2012influences}. As a result, we observe a key failure mode, \emph{tonal speech embedding collapse} \cite{feng2024towards, jiang2022promptbertimprovingbertsentence, jing2021understanding}, where different tones of the same base word become overly similar. In Hmong, for example, ``lia\textcolor{blue}{b}'' and ``lia\textcolor{blue}{s}'' share segmental content but carry different tones and encode different meanings (see Appendix~\ref{app:hmong_rpa}). In practice, these \emph{speaker-invariant pairs} and \emph{tone-variant pairs} overlap in embedding space, undermining both speaker invariance and tone awareness (Figure \ref{fig:intro_failure}). For low-resource tonal languages like Hmong, generic fine-tuning of multilingual speech encoders is brittle and prone to exploiting speaker cues \cite{garellek2023phonetics, lewis2012building}. Downstream ASR further complicates this trade-off: phone level consistency can weaken tonal contrast, while stronger tone separation can destabilize decoding \cite{shen2024encoding}.

\begin{figure*}[t!]
    \centering
    \includegraphics[width=\linewidth]{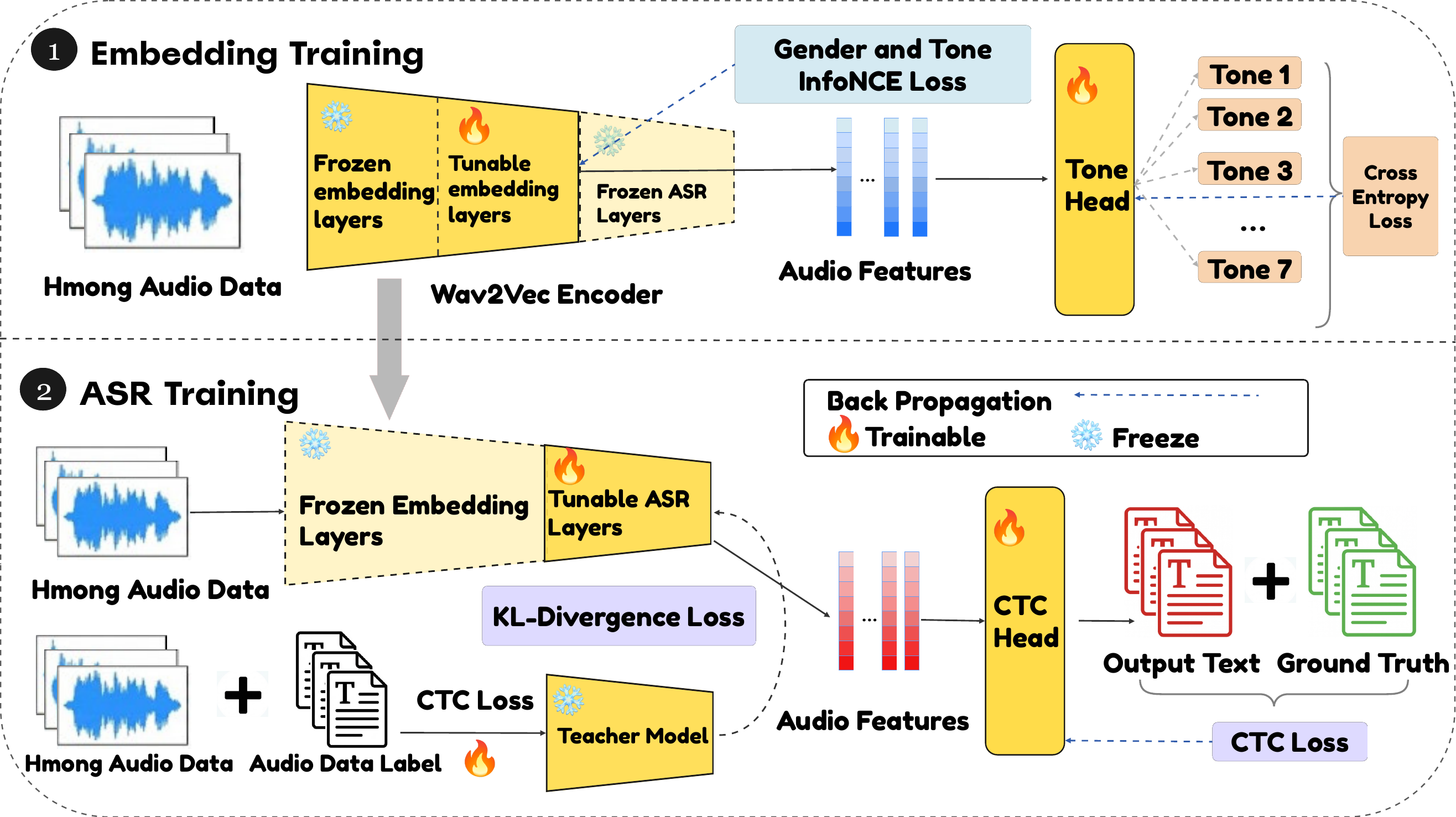}
    \vspace{-20pt}
    \caption{SITA in Two Stages. Stage 1: Speaker-invariant and Tone-sensitive Representation Learning. Stage 2: CTC Fine-tuning with Knowledge Distillation.}
    \label{fig:pipeline}
    \vspace{-15pt}
\end{figure*}

To address this, we propose \method (Speaker-Invariant yet Tone-Aware adaptation), a lightweight plug-in recipe that targets the tension between tone sensitivity and speaker invariance in tonal language recognition.
As shown in Figure~\ref{fig:pipeline}, \method proceeds in two stages. Stage 1 trains mid-layer embeddings via cross-speaker contrastive learning~\cite{Meng_2018} with explicit tone repulsion for same-base, different-tone pairs. Stage 2 fine-tunes the upper layers with CTC and ASR distillation to preserve recognition utility~\cite{tian2022knowledge, huang2018knowledge}, while lower layers remain frozen for lightweight adaptation~\cite{vanderreydt2023parameter, inoue2024elp}.

We evaluate \method on cross-speaker word retrieval, tone geometry, and ASR. Off-the-shelf multilingual encoders perform poorly here and can degrade further under na\"ive adaptation. In contrast, our parameter-efficient lightweight adaptation substantially improves retrieval, achieving an average Top-1 of $0.611$, and reduces tone collapse by increasing hard-negative cosine distance from about $0.01$--$0.08$ to $0.675$ while preserving within-tone similarity at $0.80$.
Our contributions are:
\begin{itemize}[leftmargin=13pt]
    \item We identify tonal speech embedding collapse as a key low-resource failure mode and quantify it across multilingual baselines, highlighting the invariance--awareness tension.
    \item We propose \method, a low-resource tonal adaptation recipe that composes cross-speaker contrastive, tone-repulsive, and ASR-distillation objectives in a staged schedule. We show it produces a better retrieval, ASR, and tone-geometry trade-off than label-aware and semi-supervised baselines under matched supervision.
    \item We validate \method with extensive experiments on Hmong, including unseen-speaker generalization, ablations, and recipe transfer to Mandarin as a controlled second tonal language.
\end{itemize}

\section{Related Work}
\label{sec:related_work}

\noindent\textbf{Multilingual Self-Supervised Pretraining.}
\method starts from a pretrained multilingual encoder, so we begin there. Self-supervised encoders such as wav2vec~2.0~\cite{baevski2020wav2vec20frameworkselfsupervised}, HuBERT~\cite{hsu2021hubertselfsupervisedspeechrepresentation}, and WavLM~\cite{Chen_2022}, with multilingual variants such as XLS-R~\cite{babu2021xlsrselfsupervisedcrosslingualspeech} and XLSR~\cite{conneau2020unsupervisedcrosslingualrepresentationlearning}, are the standard base for low-resource adaptation, usually with parameter-efficient tuning~\cite{houlsby2019parameterefficienttransferlearningnlp, hu2021loralowrankadaptationlarge, thomas2022efficientadaptertransferselfsupervised}. Beyond human speech, learned audio representations have enabled label-efficient bioacoustic detection through masked pretraining and self-training~\cite{xu2026mastlabelefficientrobustgeneralizable}, while decodable latent representations can support the generation of plausible ancestral vocalizations~\cite{xu2026generatingunheardphylogenyguidedlatent}. General-purpose speech pretraining objectives, however, do not explicitly state which variation to suppress and which to keep discriminative~\cite{feng2024towards, koenecke2020racial}, so low-resource fine-tuning can entangle speaker cues with lexical identity~\cite{Meng_2018}. \method keeps this backbone and this parameter-efficient setting, and adds the missing specification by suppressing speaker gender while preserving lexical tone.

\noindent\textbf{Supervised and Label-Aware Representation Learning.}
A second line injects supervision into representation learning. UniSpeech~\cite{wang2021unispeech} unifies supervised CTC with self-supervised contrastive learning for cross-lingual transfer, and LASR~\cite{vashishth2023label} uses utterance-level attribute labels to shape the space for language identification. Both therefore carry richer supervision than generic pretraining, yet both fold it into one objective aimed at general transferability. Tone collapse is a different target, because tone variants of a base word merge exactly when the model is pushed toward speaker invariance, and one objective cannot separate these two pressures. \method uses the same supervision but arranges it differently, introducing no new backbone or loss and instead staging existing objectives so that embedding shaping precedes recognition fine-tuning.

\noindent\textbf{Speaker Invariance and Disentanglement.}
Closest to our speaker-invariance term is work that removes nuisance variation directly, through adversarial training with gradient reversal~\citep{Meng_2018, adi2019reverse}, through disentanglement of speaker and content latents~\citep{hsu2017unsupervised}, or through contrastive and metric-learning objectives~\citep{oord2019representationlearningcontrastivepredictive, jiang2020speech, khosla2020supervised}. These formulations, however, state only what to suppress. In a tonal language that is not enough, because pitch carries both speaker identity and lexical tone, so stripping pitch variation to hide the speaker also erases the contrast between words. \method therefore pairs cross-speaker attraction with explicit repulsion between tone variants of one base word, stating both pressures separately instead of suppressing the speaker alone.

\noindent\textbf{Tone in Speech Representations.}
Finally, layer-wise analyses show that self-supervised models hold suprasegmental information at intermediate depths and that ASR-oriented fine-tuning shifts it~\citep{shen2024encoding, de2024layer}. \method builds on this directly, shaping an intermediate layer for tone and leaving the layers above it for recognition rather than asking one layer to do both. Because such a constraint can still destabilize decoding, we adopt teacher-student distillation for CTC, where blank-dominated posteriors make na\"ive KL matching hard~\citep{tian2022knowledge, yoon2023network}. Unlike methods that repair tonal errors after decoding, \method acts inside the encoder and reshapes what the CTC head consumes.

\section{Hmong Audio Dataset}
\label{sec:data}
We collect a Hmong word-level corpus, targeting downstream tasks such as the Word Recognition Testing (WRT) \cite{Lor2022ValidatingHmongWRT}.

\subsection{Data Collection}
Hmong speech data is collected from an ongoing WRT study with adult native speakers, under a standardized protocol with IRB-approved consent. In the testing, participants read and recorded four curated word lists that included all seven lexical tones for each word, totaling 1,400 words per participant. Speakers are instructed to produce each word in a natural voice with 3 second pause between consecutive words to ease segmentation. Appendix~\ref{app:data_stats} shows the full dataset statistics.


\subsection{Preprocessing and Augmentation}
\label{sec:data_augmentation}
We segment word-level clips with silence-based VAD~\cite{sharma2022comprehensive} and align each segment to its canonical lexical form using session metadata, so that every segment carries a lexical label, a tone label, and speaker metadata.
To reduce overfitting to speaker-specific acoustics we augment the training set in two ways. We first apply lightweight waveform perturbations such as additive noise, time stretching, and gain jitter~\cite{ko15_interspeech}. Pitch shifting is the one perturbation that is not lexically neutral in a tonal language, so we disable it for every tone-related objective and for ASR training. We then generate cross-speaker variants of each lexical item with FreeVC~\cite{li2022freevchighqualitytextfreeoneshot}, a content-preserving text-free one-shot voice conversion model, which supplies the cross-speaker positives used by our speaker-invariant loss. Voice conversion draws only on training speakers, and we audit every converted clip for signal-to-noise ratio, spectral similarity, and duration drift. Appendix~\ref{app:preprocessing_augmentation} gives the full protocol.

\section{Method}
\label{sec:method}
\method adapts a pretrained multilingual speech encoder in two stages, without introducing a new SSL backbone or contrastive loss, to learn speaker-invariant yet tone-aware representations while retaining ASR capability (Figure~\ref{fig:pipeline}). 

\subsection{Problem Definition} \label{sec:problem}
Let $\gD = \{(x_i, y_i, t_i, g_i)\}_{i=1}^{M}$ be a word-level tonal speech dataset, where $x_i$ is a waveform segment holding one spoken token, $y_i\in\mathcal{Y}$ is its transcript, $t_i\in\{1, \dots, T\}$ is its tone label, and $g_i \in \mathcal{G}$ is a demographic attribute such as gender or age. Given a pretrained multilingual speech model $f_{\theta_0}$, we treat the bottom $L$ layers as speech embedding layers and the layers above $L$ as ASR layers, so that $\vz_i = f_\theta (x_i; L)$ is an $\ell_2$-normalized embedding and forwarding $\vz_i$ through the remaining layers yields $p_{\theta}(y_i|x_i)$. We measure similarity by $\mathrm{sim}(\vu,\vv)=\vu^\top \vv$ and distance by $1-\mathrm{sim}(\vu, \vv)$.

We seek adapted parameters $\theta$ whose representations satisfy two requirements. \textbf{Speaker invariance} asks that tokens of the same lexical item stay close under demographic shift, so $\vz_i \approx \vz_j$ when $y_i = y_j$ and $g_i\neq g_j$. \textbf{Tone awareness} asks that lexical pairs differing only by tone stay well separated, so $\vz_i \not\approx \vz_j$ when \texttt{base}($x_i$) $=$ \texttt{base}($x_j$) and $t_i\neq t_j$. The representation must also retain recognition-friendly structure for ASR.


\subsection{\method} 
SITA consists of two stages. Single-stage joint optimization of all losses tends to either undermine tone separation when ASR dominates the gradient, or destabilize recognition when contrastive losses dominate. We therefore decouple the optimization: Stage 1 first shapes a tone-structured embedding space using two complementary loss terms, then Stage 2 restores recognition capability for ASR via CTC fine-tuning while freezing the Stage 1 layers, optionally regularized by knowledge distillation.

\subsubsection*{Stage 1: Speech Representation Learning.}

\noindent\textbf{Speaker-invariant Loss.} For an anchor segment $x_i$, we construct one \emph{cross-speaker} positive $x_i^{+}$ that is another recording of the \emph{same lexical item} spoken by a speaker of the opposite gender (see speaker-diversified augmentation in \S\ref{sec:data_augmentation}). We instantiate the demographic attribute as gender for simplicity, though the loss generalizes to other attributes such as age. We sample $N$ negatives $\{x_{i,n}^{-}\}_{n=1}^{N}$ from \emph{different lexical items} (thus carrying different tones). We define the per-anchor InfoNCE loss $\ell_1$ as the negative log-likelihood of selecting the cross-speaker positive among one positive and $N$ negatives. Specifically, we compute the similarities scaled by a temperature $\tau_g$ for the positive and negative pairs and the loss as: 
\setlength{\jot}{0.15em}
\begin{align}
s_i^{+} = \frac{\operatorname{sim}(\vz_i, \vz_i^{+})}{\tau_g},
\; s_{i,n}^{-} = \frac{\operatorname{sim}(\vz_i, \vz_{i,n}^{-})}{\tau_g},
\label{eq:gender_logits}
\\
\ell_1 (x_i) = -\log
\frac{\exp(s_i^{+})}
{\exp(s_i^{+}) + \sum_{n=1}^{N}\exp(s_{i,n}^{-})}.
\label{eq:gender_loss_i}
\end{align}

\noindent Aggregating all segments $x_i$ from $\mathcal{D}$ of size $M$, we optimize the InfoNCE objective:
\begin{align}
\mathcal{L}_{\text{speaker}}(\mathcal{D}) = \frac{1}{|\gD|}\sum_{x_i\in\gD}\ell_1 (x_i) .
\label{eq:gender_loss}
\end{align}
This objective treats gender as within-class nuisance variation: it pulls together embeddings of the same lexical item spoken by different speaker genders while pushing away different lexical items.

\noindent\textbf{Tone-aware Loss.}
Using the lexical-preserving acoustic augmentation (\S\ref{sec:data_augmentation}), we additionally construct three sets for each anchor $x_i$: (i) a positive set $\gP(i)$ of waveform segments with the same word and tone, (ii) a set of hard negatives $\gH(i)$ with the same word but different tones, and (iii) a set of soft negatives $\gS(i)$ from different words.
For each anchor segment $x_i$, we compute the similarity scaled by a temperature $\tau_t$ with all segments in the three constructed sets, and the normalization term $Z_i$ as:
\setlength{\jot}{0.15em}
\begin{align}
s_{ij} &= \frac{\operatorname{sim}(\vz_i, \vz_j)}{\tau_t},~\text{for}~ j \in \gP(i) \cup \gH(i) \cup \gS(i), \nonumber\\
Z_i &= \sum_{j \in \gP(i)\cup \gH(i) \cup \gS(i)} \exp(s_{ij}),
\label{eq:tone_partition}
\end{align}
For each anchor $x_i$ with multiple tone positives, the per-sample loss $\ell_2$ is
\setlength{\jot}{0.15em}
\begin{align}
\ell_2 (x_i)
=
-\frac{1}{|\gP(i)|}
\sum_{j \in \gP(i)}
\log
\frac{\exp(s_{ij})}{Z_i}.
\label{eq:tone_loss_i}
\end{align}

We also add a tone classifier to steer the embedding toward tone-discriminative structure. Specifically, we add a linear layer for predicting the tone label from the embedding, i.e., $p_\phi(t\mid \vz)=\mathrm{softmax}(\mW \vz+b)$ with tunable parameters $\phi = \{\mW, b\}$. Here, we use the standard cross-entropy loss: $\ell_{3}(x_i, t_i) = -\log p_\phi(t_i\mid \vz_i)$.

Lastly, the tone-aware loss is then computed as the weight sum of $\ell_2$ and $\ell_3$ over all $(x_i, t_i)$ in the dataset $\gD$ of size $M$:
\setlength{\jot}{0.15em}
\begin{align}  \nonumber
\mathcal{L}_{\text{tone}} (\gD)
=
\frac{1}{|\gD|}
\sum_{(x_i, t_i)\in \gD}
\left( \ell_2 (x_i) + \lambda \ell_3 (x_i, t_i) \right). %
\end{align}

\noindent\textbf{Stage 1 Objective.}
We jointly optimize the weighted sum of the two loss terms in Stage 1. 
\begin{align}
\nonumber
\mathcal{L}_{\text{Stage-1}}(\gD)=\alpha\,\mathcal{L}_{\text{speaker}}(\gD) +(1-\alpha)\mathcal{L}_{\text{tone}}(\gD).
\end{align}
For parameter-efficient fine-tuning, we update only the layers up to the embedding extraction layer $L$ and reserve the upper layers for Stage 2 ASR (Figure~\ref{fig:pipeline}).

\subsubsection*{Stage 2: ASR Fine-tuning}
To preserve the embeddings learned from Stage 1, we freeze the bottom $L$ layers and only update the layers above $L$ together with the CTC projection head. We regularize the student ASR with a frozen ASR teacher via knowledge distillation.

\noindent\textbf{CTC Loss.}
Given encoder frame representations $\{\vh_t\}_{t=1}^T$, we compute per-frame logits $\vu_t = \mW\vh_t$ with a linear CTC head $\mW$ and posteriors $p_{\theta}(\pi_t \mid x)=\mathrm{softmax}(\vu_t)_{\pi_t}$ over the vocabulary $\mathcal{V}\cup\{\varnothing\}$. CTC defines the sequence likelihood by summing over all alignments $\pi$ that collapse to $y$ under $\mathcal{B}(\cdot)$:
\begin{align}
p_{\theta}(y \mid x)
&= \sum_{\pi:\,\mathcal{B}(\pi)=y}\ \prod_{t=1}^{T} p_{\theta}(\pi_t \mid x).
\label{eq:ctc_marginal}
\end{align}
We minimize the negative log-likelihood over $\gD$:
\begin{align}
\mathcal{L}_{\text{CTC}}(\gD)
&= -\frac{1}{|\gD|} \sum_{(x_i, y_i)\in \gD} \log p_{\theta}(y_i \mid x_i).
\label{eq:ctc_loss_def}
\end{align}



\noindent\textbf{Knowledge Distillation (KD).} 
We use KD to keep the student’s frame-level posterior distribution close to a stronger ASR teacher parameterized by $\varphi$, improving training stability and preventing the encoder from drifting away from recognition-friendly representations. Let $\vu_{t}'$ and $\vu_{t}$ be the teacher and student CTC logits at frame $t$ respectively for a segment $x_i$.
With a temperature $\tau_\text{kd}>0$, we compute the softened teacher and student posteriors as $\tilde{p}_\varphi(\pi_t\mid x_i)=\operatorname{softmax}(\vu'_{t}/\tau_{\text{kd}})$ and
$\tilde{p}_\theta(\pi_t\mid x_i)=\operatorname{softmax}(\vu_{t}/\tau_{\text{kd}})$.
We compute the frame-wise KL divergence between the teacher and student posteriors for all segment $x_i\in \gD$.
\begin{align} \nonumber
\mathcal{L}_{\text{KD}} (\gD)
= \frac{1}{|\gD|} \sum_{x_i\in \gD} \operatorname{KL}(\tilde{p}_\varphi(\pi\mid x_i) || \tilde{p}_\theta(\pi\mid x_i)).
\end{align}

\noindent\textbf{Stage-2 Objective.}
We minimize a weighted sum of the CTC and KD loss terms with a weight $\delta$.
\begin{align} \nonumber
\mathcal{L}_\text{Stage-2} (\gD)
=
\delta\,\mathcal{L}_{\text{CTC}}(\gD)
+
(1-\delta)\mathcal{L}_{\text{KD}}(\gD).
\end{align}

\section{Experiment Setup}




\noindent\textbf{Datasets and Setup.}
We evaluate primarily on the Hmong word-level corpus from \S\ref{sec:data}, and additionally on \emph{Tone Perfect},\footnote{\url{https://tone.lib.msu.edu/}} a Mandarin corpus of monosyllables across four tones, to test whether the tone objectives generalize beyond Hmong. Statistics, splits, and hyperparameters are in Appendices~\ref{app:data_stats} and \ref{app:training_settings}. The encoder has 24 Transformer blocks over a convolutional feature extractor, with a CTC head for ASR. We report embedding extraction layers $\ell \in \{19, 21\}$, selected by layer-wise probing (Appendix~\ref{app:feature_layer_selection}). The \emph{embedding extraction layer} is the block whose pooled output we use as the word embedding, not the block being updated. Stage 1 updates blocks 13 through that layer and freezes the rest, and Stage 2 freezes every block up to and including it while updating the remaining upper blocks and the CTC head. Appendix~\ref{app:layer_indexing_freezing} states the indexing and freezing convention in full, and Appendix~\ref{app:pooling} gives the pooling used to form word embeddings.

\noindent\textbf{Evaluation.}
We evaluate three tasks and define all metrics in Appendix~\ref{app:metrics}. \textbf{Cross-gender word retrieval}~\cite{kamper2015unsupervised, carlin2011rapid} ranks opposite-gender utterances by cosine similarity given a query and reports Top-1 and Top-5 accuracy. \textbf{Tone geometry} reports cosine similarity on positives, which share word and tone, and cosine distance on hard negatives, which share the base word but differ in tone, and on soft negatives, which are different words. \textbf{ASR}~\cite{morris2004and} decodes with CTC beam search and reports WER and CER.

\noindent\textbf{Baselines.}
All baselines start from the same multilingual XLS-R encoder~\citep{babu2021xlsrselfsupervisedcrosslingualspeech}, and we pool encoder hidden states whenever a baseline needs retrieval or tone metrics. The \emph{off-the-shelf or data-only} group sees no Hmong-specific labels and contains frozen \textbf{XLS-R}, \textbf{SSL Adaptation} that continues wav2vec~2.0-style self-supervised pretraining on Hmong, \textbf{ASR Adaptation} fine-tuned with CTC, and zero-shot \textbf{Whisper}~\citep{radford2022robustspeechrecognitionlargescale} and \textbf{Omnilingual ASR}~\citep{omnilingualasrteam2025omnilingualasropensourcemultilingual}. The \emph{matched-supervision} group sees the same gender, tone, and word labels as \method. \textbf{Speaker-Adversarial (GRL)}~\citep{Meng_2018, adi2019reverse} replaces $\mathcal{L}_{\text{speaker}}$ with a gradient-reversal gender classifier, \textbf{LASR}~\citep{vashishth2023label} is label-aware representation learning without our staged schedule, and \textbf{UniSpeech}~\citep{wang2021unispeech} is a semi-supervised framework adapted to our supervision regime.
\section{Results and Analysis}

\subsection{Task Evaluation}
\label{sec:results_retrieval}

\begin{figure}[t]
\centering
\vspace{-2mm}
\includegraphics[width=\linewidth,keepaspectratio]{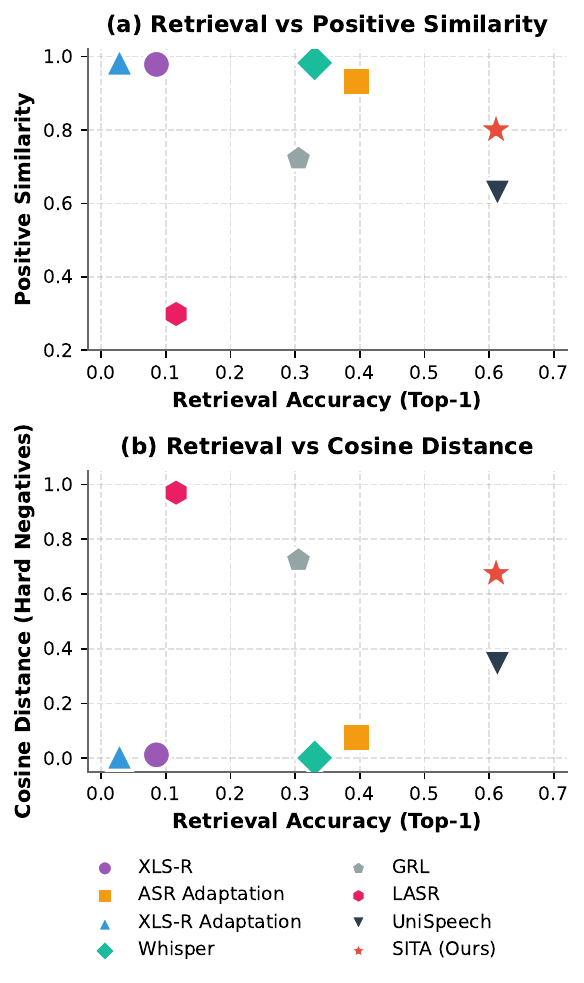}
\vspace{-20pt}
\caption{Trade-off between demographic robustness and tone geometry on Hmong. We plot Avg. Top-1 cross-gender retrieval against (a) positive similarity (same word, same tone) and (b) hard-negative cosine distance (same word, different tone). Baselines exhibit tone collapse while SITA achieves strong retrieval with substantial cross-tone separation.}
\label{fig:hmong_tone_tradeoff}
\vspace{-15pt}
\end{figure}

\textbf{SITA Learns Gender-Invariant Lexical Embeddings.}
Tone in Hmong is pitch-encoded and male and female pitch ranges differ substantially, so a model that conflates speaker identity with tone fails disproportionately on cross-gender retrieval. Table~\ref{tab:hmong_retrieval_main} reports Top-1 and Top-5 accuracy for female queries against a male gallery (F$\rightarrow$M) and male queries against a female gallery (M$\rightarrow$F). Off-the-shelf XLS-R and SSL Adaptation are near chance or worse, while ASR Adaptation and Whisper are stronger but still limited.
Among matched-supervision baselines, GRL~\cite{Meng_2018} is unstable across 
directions, LASR~\cite{vashishth2023label} is highly asymmetric, and 
UniSpeech~\cite{wang2021unispeech}, while competitive on retrieval, trades off 
weaker tone separation and worse ASR 
(Figure~\ref{fig:hmong_tone_tradeoff}, Table~\ref{tab:hmong_asr_main}).
\method achieves $0.6286$/$0.5929$ Top-1 and $0.9286$/$0.8893$ Top-5 on 
F$\rightarrow$M and M$\rightarrow$F, with the most consistent balance across 
retrieval, tone geometry, and ASR.
Full results are in Appendix~\ref{app:full_tables_subsec}.
\begin{table}[h]
\centering
\setlength{\tabcolsep}{3.5pt} 
\renewcommand{\arraystretch}{0.95} 
\vspace{-6pt}
\caption{Cross-gender word retrieval on Hmong for female queries against a male gallery (F$\rightarrow$M) and male queries against a female gallery (M$\rightarrow$F). Off-the-shelf baselines see no Hmong-specific supervision. Matched-supervision baselines use the same labels as \method, namely gender, tone, and lexical identity. \method is its best retrieval configuration, embedding extraction layer 21 with frozen adaptation.}
\vspace{-6pt}
\label{tab:hmong_retrieval_main}
\resizebox{0.48\textwidth}{!}{
\begin{tabular}{lcccc}
\toprule
\multirow{2}{*}{\textbf{Model}} & \multicolumn{2}{c}{\textbf{F$\rightarrow$M}} & \multicolumn{2}{c}{\textbf{M$\rightarrow$F}} \\
\cmidrule(lr){2-3}\cmidrule(lr){4-5}
& Top-1 & Top-5 & Top-1 & Top-5 \\
\midrule
\multicolumn{5}{l}{\emph{Off-the-shelf / data-only baselines}} \\
XLS-R  & 0.0714 & 0.2143 & 0.1000 & 0.2429 \\
XLS-R Adaptation & 0.0286 & 0.0857 & 0.0286 & 0.1000 \\
ASR Adaptation & 0.3286 & 0.6714 & 0.4607 & 0.7786 \\
Whisper (Large-v3) & 0.2857 & 0.6143 & 0.3750 & 0.7036 \\
\midrule
\multicolumn{5}{l}{\emph{Matched-supervision baselines}} \\
GRL \cite{Meng_2018} & 0.3714 & 0.7286 & 0.2393 & 0.6107 \\
LASR \cite{vashishth2023label} & 0.2286 & 0.6857 & 0.0036 & 0.0143 \\
UniSpeech \cite{wang2021unispeech} & \textbf{0.6827} & \textbf{0.9429} & 0.5429 & 0.8179 \\
\midrule
\textbf{SITA} & 0.6286 & 0.9286 & \textbf{0.5929} & \textbf{0.8893} \\
\bottomrule
\end{tabular}}
\vspace{-13pt}
\end{table}



\noindent\textbf{SITA Preserves Tone Structure without Sacrificing Cross-Gender Retrieval.}
\label{sec:results_tone}
Figure~\ref{fig:hmong_tone_tradeoff} plots Top-1 cross-gender retrieval against positive similarity and against hard-negative cosine distance, and it shows that the methods with high retrieval accuracy are also the ones that avoid \emph{tone collapse}. In Fig.~\ref{fig:hmong_tone_tradeoff}(a), the lower positive similarity of SITA reflects reduced global collapse, since most baselines stay highly similar even for mismatched pairs while SITA makes similarity selective. Baselines also fail to separate different tones of the same word, with hard-negative cosine distance between $0.01$ and $0.08$, which coincides with weak cross-gender retrieval. SITA instead keeps within-tone cohesion and raises hard-negative distance to $0.675$ at Top-1 $0.611$, the best trade-off between the two axes. The matched-supervision baselines fail in two different ways. LASR pushes hard-negative distance to $0.970$ but collapses retrieval to $0.116$, and UniSpeech reaches retrieval $0.613$ with weak tone separation at $0.343$. GRL reaches hard-negative distance $0.723$ yet still underperforms on retrieval, which suggests that adversarial invariance alone can distort lexical structure without producing robust word-level embeddings. Appendices~\ref{app:tone_cls}, \ref{app:sim_eval}, and~\ref{app:per_tone} give additional diagnostics on tone classification, similarity scoring robustness, and per-tone breakdowns.


\begin{table}[h]
\centering
\setlength{\tabcolsep}{3.5pt} 
\renewcommand{\arraystretch}{0.95}
\vspace{-3pt}
\caption{ASR on Hmong word+tone, reported as a stability check that Stage 2 preserves usable recognition structure. \method is its best ASR configuration, embedding extraction layer 19 with frozen adaptation and KD. Lower CER and WER are better.}
\vspace{-8pt}
\label{tab:hmong_asr_main}
\resizebox{0.35\textwidth}{!}{
\begin{tabular}{lcc}
\toprule
\textbf{Model} & \textbf{CER} & \textbf{WER} \\
\midrule
Whisper (Large-v3) & 1.1181 & 1.1078 \\
OmnilingualASR 7B & 0.9994 & 1.0000 \\
\midrule
ASR Adaptation & 0.1835 & 0.4610 \\
\midrule
SITA Stage 2 & 0.1985 & 0.5115 \\
\bottomrule
\end{tabular}}
\vspace{-15pt}
\end{table}

\noindent\textbf{Stage 2 Preserves Recognition-Relevant Structure.}
Stage 2 is a stability check. As shown in Table~\ref{tab:hmong_asr_main}, existing multilingual ASR models recognize Hmong poorly, which motivates targeted adaptation. The ASR Adaptation baseline is XLS-R fine-tuned with CTC, and it serves both as the Stage 2 teacher and as the low-resource reference point. Relative to it at WER $0.4610$, \method Stage 2 increases WER slightly to $0.5115$ while enabling the representation gains reported above. Appendix~\ref{app:full_tables} reports full results.

\begin{figure*}[t]
\centering
\vspace{-2mm}
\includegraphics[width=\linewidth,height=0.30\textheight,keepaspectratio]{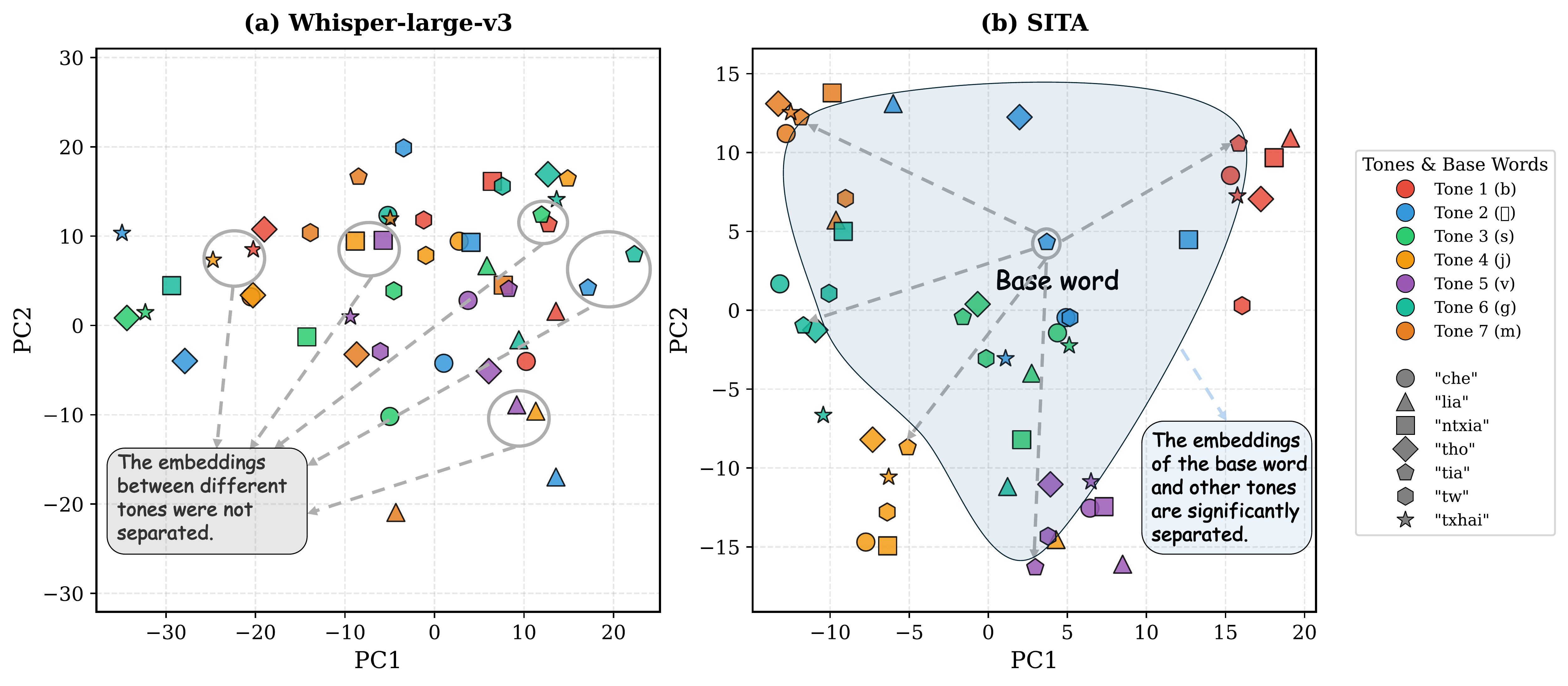}
\vspace{-20pt}
\caption{
We visualize token embeddings using PCA (2D) computed from $\ell_2$-normalized pooled encoder representations. Points denote word tokens, colors are tones, and marker shapes are base words. Dashed lines connect tone variants within each base word. Whisper shows tone collapse (within-word tones overlap), whereas SITA separates tone variants into a more tone-stratified geometry, aligning with improved hard-negative separation and retrieval. }
\vspace{-14pt}
\label{fig:case_study}
\end{figure*}

\noindent\textbf{SITA Reduces Tone Collapse while Preserving Within-Tone Cohesion.} In Figure \ref{fig:case_study}, we show the PCA results of different tones' embeddings for Whisper and SITA. On test tokens that share the same base word across different tones, Whisper embeddings largely collapse into overlapping clusters. 
In contrast, SITA produces a more tone-stratified geometry while keeping same-tone tokens compact, which aligns with its stronger retrieval performance and larger hard-negative separation. 
This qualitative pattern is consistent with the decoding behavior of the two models on the same tokens, where Whisper produces frequent wrong-word errors while SITA stays within the correct base word and errs only on tone. Appendix~\ref{app:case_study} reports this case study in full.

\begin{table*}[t]
\centering
\small
\setlength{\tabcolsep}{3.2pt}
\renewcommand{\arraystretch}{0.95}
\caption{Component ablations for \textbf{SITA}. \textsc{Embed.\ Layer} is the encoder layer whose pooled output is used as the word embedding for retrieval and tone analyses. \textsc{Stage 1 freeze} indicates freezing the bottom 12 layers. \textsc{Stage-2 \#Upd.}\ counts trainable encoder layers (0 = head-only). Retrieval: Top-1/Top-5 averaged over the F$\rightarrow$M and M$\rightarrow$F directions. ASR: CER/WER.}
\vspace{-8pt}
\label{tab:ablation_components}
\begin{tabular}{ccccc|cccc}
\toprule
\multicolumn{5}{c|}{\textbf{Components}} & \multicolumn{4}{c}{\textbf{Metrics}} \\
\cmidrule(lr){1-5}\cmidrule(lr){6-9}
\textsc{Staged}
& \textsc{Embed.\ Layer}
& \textsc{Stage 1 Freeze}
& \textsc{Stage 2~\#Upd. layers}
& \textsc{KD}
& Avg T1 & Avg T5 & CER & WER \\
\midrule
\cmark & 19 & \xmark & 5 &  \xmark & 0.5215 & 0.8786 & 0.2140 & 0.5390 \\
\cmark & 19 & \cmark & 5 &  \xmark & 0.5465 & 0.9000 & 0.2094 & 0.5344 \\
\cmark & 19 & \cmark & 5 &  \cmark & 0.5465 & 0.9000  & 0.1985 & 0.5115 \\
\cmark & 21 & \cmark & 3 &  \xmark & 0.6108 & 0.9090 & 0.2261 & 0.5505 \\
\cmark & 24 & \cmark & 12 &  \xmark & 0.3661 & 0.7340 & 0.1968 & 0.4771 \\
\cmark & 24 & \cmark & 0 &  \xmark & 0.5679 & 0.8482 & 0.2457 & 0.5665 \\
\midrule
\xmark & 21 & \cmark & \_ &  \xmark & 0.6893 & 0.9304 & 0.3421 & 0.7637 \\
\xmark & 24 & \cmark & \_ &  \xmark & 0.7072 & 0.9598 & 0.3538 & 0.7816 \\
\bottomrule
\end{tabular}
\vspace{-10pt}
\end{table*}

\subsection{Ablation Studies}
\label{sec:results_ablation}

Table~\ref{tab:ablation_components} ablates the core design choices of \method along two axes, representation quality measured by cross-gender lexical retrieval and ASR stability measured by CER and WER.
Main-paper tables report the best \method configuration per metric. Stage 2 freezes the Stage 1 layers, so KD changes ASR but not retrieval. Further studies are in the appendix. Appendix~\ref{app:freevc_ablation} removes
voice conversion, Appendix~\ref{app:pooling_ablation} varies pooling, and
Appendix~\ref{app:tone_margin} swaps the InfoNCE tone objective for a
margin-based one. Appendix~\ref{app:wavlm_backbone} repeats the recipe on a
WavLM backbone, Appendix~\ref{app:non_tonal_probe} probes non-tonal English
keywords as a regression check, and Appendix~\ref{app:sentence_probe} extends
the evaluation to four-word pseudo-sentences.

\noindent\textbf{Staged Training is Important.}
Single-stage optimization of all objectives reaches Top-1 accuracy up to $0.7072$ but produces much weaker ASR. The staged recipe gives a substantially better trade-off between retrieval and ASR. Our 19-layer setting reaches Avg Top-1 $0.547$ and Top-5 $0.900$ at WER $0.534$, and the best KD variant improves ASR to $0.511$.

\noindent\textbf{The Embedding Extraction Layer Controls the Trade-off.}
When the embedding extraction layer is the last block, updating many encoder layers in Stage 2 yields strong ASR at WER $0.477$ but severely degrades the embedding space, whereas restricting Stage 2 to head-only training preserves retrieval at $0.568$ and weakens ASR to WER $0.567$. Intermediate embedding extraction layers 19 and 21 balance the two, giving strong retrieval with substantially better ASR than single-stage training.

\noindent\textbf{Freezing Helps Retrieval, While KD Improves ASR.}
Freezing the first 12 layers in Stage 1 gives a modest but consistent retrieval gain in the 19-layer setting, where Avg. Top-1 rises from $0.522$ to $0.546$ and ASR also improves slightly. Knowledge distillation helps ASR in the same configuration, improving WER from $0.534$ to $0.512$ while leaving retrieval unchanged, so we use it in our final system.

\begin{table}[h]
\centering
\small
\vspace{-4pt}
\caption{Unseen-speaker cross-gender retrieval on Hmong (M$\rightarrow$F), frozen adaptation. Higher is better.}
\vspace{-8pt}
\label{tab:hmong_unseen_retrieval}

\begin{adjustbox}{width=0.8\columnwidth}
\begin{tabular}{lcc}
\toprule
\textbf{Model} & \textbf{Top-1} & \textbf{Top-5} \\
\midrule
XLS-R & 0.1203 & 0.3254 \\
ASR Adaptation & 0.5453 & 0.8731 \\
SSL Adaptation & 0.0741 & 0.2142 \\
Whisper (Large-v3) & 0.2191 & 0.5206 \\
GRL & 0.4086 & 0.6787 \\
\midrule
\textbf{SITA} & \textbf{0.6870} & \textbf{0.9077} \\
\bottomrule
\end{tabular}
\end{adjustbox}
\vspace{-6pt}
\end{table}

\subsection{Generalization to Unseen Speakers}
\label{sec:unseen_speakers}

We evaluate speaker shift with a held-out split containing one unseen male speaker, who supplies the queries while the seen female speakers form the gallery. The reverse direction is not available here. In Table~\ref{tab:hmong_unseen_retrieval}, ASR Adaptation is the strongest baseline at $0.5453$ Top-1 and $0.8731$ Top-5, and \method improves these to $0.6870$ and $0.9077$, so the recipe stays robust when the speaker is unseen. Appendix~\ref{app:unseen_full} reports additional configurations.

\subsection{Generalization to Other Tonal Languages}

\begin{table}[t]
\centering
\small
\caption{Cross-gender word retrieval accuracy on Mandarin (Male$\rightarrow$Female). We report Top-1 and Top-5 accuracy for the multilingual baseline, a Mandarin ASR teacher, and \method.}
\vspace{-2pt}
\label{tab:zh_retrieval}
\begin{tabular}{lcc}
\toprule
\textbf{Model} & \textbf{Top-1} & \textbf{Top-5} \\
\midrule
XLS-R & 0.2457 & 0.4768 \\
ASR Adaptation & 0.9622 & 0.9957 \\
SSL Adaptation & 0.0640 & 0.1780  \\
Whisper (Large-v3) & 0.8476 & 0.9738  \\
\midrule
\textbf{SITA} & \textbf{0.9927} & \textbf{0.9994} \\
\bottomrule
\end{tabular}
\vspace{-10pt}
\end{table}
We apply the same architecture, objectives, and hyperparameters to Mandarin on the Tone Perfect corpus, keeping the XLS-R initialization and swapping only the training data and tone inventory. Table~\ref{tab:zh_retrieval} shows that off-the-shelf XLS-R has limited cross-gender robustness while Mandarin ASR Adaptation and Whisper are strong baselines. SITA reaches near-ceiling Male$\rightarrow$Female retrieval at Top-1 $\approx$$0.99$, so enforcing gender invariance does not degrade lexical discrimination under a different tonal system. We see the same tone geometry as in Hmong, where baselines keep different tones of a syllable overly close while our objective increases cross-tone separation without losing within-tone similarity. Full results are in Appendix~\ref{app:mandarin_full}.

\section{Conclusion}

We address low-resource tonal-language adaptation, where embeddings should be speaker-invariant yet tone-aware. Stage 1 learns gender-invariant and tone-sensitive embeddings from cross-gender contrastive supervision and tone repulsion, and Stage 2 restores ASR by CTC fine-tuning with optional distillation. On Hmong, \method sharpens tone separation and improves cross-gender retrieval over multilingual and ASR baselines, and it outperforms GRL, LASR, and UniSpeech on the trade-off across retrieval, tone geometry, and ASR under matched supervision. The recipe transfers to Mandarin and to a WavLM backbone, making \method a general plug-in for low-resource tonal adaptation.


\section*{Limitations}

Our evaluation of SITA is conducted on a curated word-level corpus with limited coverage of speakers, recording conditions, and speaking styles. As a result, our retrieval and tone-geometry metrics may not fully reflect performance on spontaneous, conversational speech or under broader domain shift.

We use word-level data because our Hmong corpus was collected as isolated word speech for a speech test, and because isolated words let us construct controlled tonal pairs that probe tone collapse most cleanly. To check that \method's properties are not narrowly word-level, we run a controlled longer-context probe in Appendix~\ref{app:sentence_probe}, where pseudo-sentences are constructed by concatenating word clips. \method is the only method whose tone awareness and speaker invariance carry over to this 4-word setting. A full evaluation on natural continuous speech, which involves coarticulation, prosody, and word-level tone alignment in long-form recordings, remains future work.

While we observe consistent gains with lightweight partially frozen adaptation, the best configuration may change with more data, different demographic distributions, or different tonal inventories. Finally, SITA exposes an inherent trade-off between tone separation and ASR accuracy. The appropriate loss weighting and freezing strategy may depend on the downstream priority, such as robust retrieval versus recognition, and may require task-specific tuning.

\section*{Ethical Considerations}

SITA targets speech technology for tonal low-resource languages that are often under-served, with potential benefits for education, accessibility tools, and language documentation/preservation. However, speech representations can encode sensitive attributes (e.g., speaker identity, gender, accent) and may enable profiling or surveillance. While our training objectives encourage invariance to nuisance variation, they do not guarantee the removal of all demographic information or prevent downstream misuse. Our evaluation is limited in speaker coverage and may miss disparate failure modes for underrepresented groups, and future work should expand demographic diversity and incorporate community-informed testing and governance. Finally, similarity or retrieval signals from embeddings should not be used as the sole basis for high-stakes decisions, and any deployment in consequential settings requires rigorous validation, oversight, and informed consent.

\section*{Acknowledgment}
Research reported in this publication was partially supported by the ICTR Pilot Awards Program at the University of Wisconsin-Madison and the National Science Foundation under Award Number IIS-2449768. The content is solely the responsibility of the authors and does not necessarily represent the official views of the National Science Foundation.

\bibliography{anthology,custom}
\bibliographystyle{acl_natbib}

\appendix

\section{Experimental Details}
\label{sec:appendix}

This appendix provides supplementary implementation details and additional results for \textbf{SITA}, omitted from the main paper for clarity.

\subsection{Hmong Tones and RPA Marking} 
\label{app:hmong_rpa} 
Hmong is a predominantly monosyllabic tonal language where lexical tone changes meaning. The Romanized Popular Alphabet (RPA) marks seven tones: high level (b), mid level (no marker), low level (s), high falling (j), mid rising (v), breathy falling (g), and low falling creaky (m). Words can share identical segments but differ only in tone, yielding distinct meanings (e.g., \textit{liab} “red” vs.\ \textit{liam} “to blame”).

\subsection{Dataset Statistics}
\label{app:data_stats}

\paragraph{Hmong (WRT) corpus.}
We use a curated word-level corpus collected under a standardized WRT protocol (\S~\ref{sec:data}).
After VAD-based segmentation and metadata alignment, the dataset contains 8,570 words from 8 speakers (3 female, 5 male), covering 1,143 unique word forms spanning 163 base words and all 7 lexical tones.
To mitigate speaker imbalance and increase within-lexeme diversity, we additionally apply voice-conversion augmentation to generate 3,600 synthetic words, yielding 12,170 total words.
The resulting audio totals 1h 41m 24s.

\paragraph{Splits.}
We construct train/test splits to ensure broad lexical coverage in the test set.
Concretely, for each base word and its tonal realizations, we place at least one token into the test split by sampling from a subset of speakers in the WRT study, continuing until all base words are represented. All remaining tokens are used for training, which gives 6,857 training tokens before augmentation.
For \textbf{unseen-speaker} evaluation (\S~\ref{sec:unseen_speakers}), we further hold out one additional male speaker as the query set, and evaluate M$\rightarrow$F retrieval against a gallery formed from the seen female speakers (Appendix~\ref{app:unseen_full}).
Unless otherwise stated, all metrics are computed on word-level segments.

\paragraph{Mandarin (Tone Perfect).}
For the Mandarin transfer experiment, we use the Tone Perfect corpus of read monosyllables with speaker and tone annotations.
Our subset contains 9,840 tokens from 6 speakers, covering 410 syllables and 4 tones.
We use one male speaker as the test set and the remaining speakers for training.

\subsection{Preprocessing and Augmentation Details}
\label{app:preprocessing_augmentation}

\paragraph{Preprocessing.}
All recordings are converted to single-channel audio at a fixed sampling rate and peak-normalized before segmentation.
We segment word-level clips with silence-based VAD~\cite{sharma2022comprehensive}, which is reliable here because the recording protocol mandates a three second pause between consecutive words.
Each segment is aligned to its canonical lexical form using session metadata and the order of the word list read by the participant.
The resulting corpus consists of clean word-level segments carrying lexical labels, tone annotations, and speaker metadata including gender, and we use the same segments for both representation learning and ASR training.
Our segmentation and alignment pipeline is implemented in our preprocessing toolkit and will be released with our codebase.

\paragraph{Acoustic perturbations.}
Data augmentation is a standard way to improve robustness under low-resource conditions, and prior work has used SpecAugment~\citep{Park_2019}, speed perturbation~\citep{ko15_interspeech}, vocal-tract-length perturbation~\citep{jaitly2013vocal}, and reverberation or noise simulation with room impulse responses~\citep{ko2017study}.
We apply lightweight waveform perturbations at training time, namely additive noise, time stretching, pitch shifting, and gain jitter.
Pitch shifting is the one operation that is not lexically neutral in a tonal language, so we disable it whenever a batch contributes to a tone-related objective and whenever it is used for ASR training.
The remaining perturbations leave the lexical label untouched while increasing channel and prosodic variability, which benefits both contrastive learning and ASR.

\paragraph{Voice conversion.}
Synthetic augmentation through text-to-speech and voice conversion has also been explored for low-resource ASR~\citep{baas22_interspeech}.
We use FreeVC~\cite{li2022freevchighqualitytextfreeoneshot}, a content-preserving text-free one-shot voice conversion model, to produce additional views of each word token under a different speaker identity, which supplies the cross-speaker positives required by $\mathcal{L}_{\text{speaker}}$.
Conversion is performed offline and only on the training split. We fix a pool of three training speakers and convert each training token to a different speaker in that pool, so no test speaker appears as either source or target and no test utterance is ever converted.
This adds 3,600 converted tokens on top of the 6,857 original training tokens, and we use them only as extra views for Stage~1 representation learning. Because a converted utterance shares its lexical label with the original, it introduces no new lexical content and cannot create near-duplicate test items.
We audit every converted clip for signal-to-noise ratio, spectral similarity, and duration drift in Appendix~\ref{app:freevc_ablation}.

\subsection{Training Settings and Hyperparameters}
\label{app:training_settings}

\paragraph{Backbone and embedding extraction.}
All models are initialized from the multilingual wav2vec-style encoder XLS-R \citep{babu2021xlsrselfsupervisedcrosslingualspeech}.
 Unless otherwise specified, token embeddings are formed by pooling Transformer hidden states (\textbf{max} pooling for retrieval; \textbf{mean} pooling for tone geometry, tone classification, and similarity analyses).(Appendix~\ref{app:pooling}). We consider two adaptation modes: \textbf{freeze}, where the bottom 12 Transformer blocks are frozen, and \textbf{non-freeze}, where all blocks are trainable. For representation evaluation, embeddings are extracted from Transformer block $\ell \in {19, 21}$, where $\ell$ denotes the block whose output is pooled as the representation interface (not the number of trainable blocks). We use 1-indexed numbering over $M=24$ blocks, using the convetion in Appendix~\ref{app:layer_indexing_freezing}. Unless otherwise noted, cross-gender retrieval uses \textbf{max} pooling and tone/similarity analyses use \textbf{mean} pooling 

\paragraph{Stage 1: Speech Representation Learning.}
We train with the joint objective
$\mathcal{L}_{\text{Stage 1}}=\alpha\mathcal{L}_{\text{speaker}}(\mathcal{D})+(1-\alpha)\mathcal{L}_{\text{tone}}(\mathcal{D})$
(\S~\ref{sec:method}).
We set $\alpha=0.5$.
For $\mathcal{L}_{\text{speaker}}$ we use temperature $\tau_g=0.07$ and sample $N=20$ negatives per anchor.
For $\mathcal{L}_{\text{tone}}$, we use $\tau_t=0.07$ and $\lambda_{\text{cls}}=1.0$.

\paragraph{Stage 2: ASR Fine-tuning with CTC and KD.}
Stage 2 minimizes
$\mathcal{L}_{\text{Stage 2}}=\delta\mathcal{L}_{\text{CTC}}(\mathcal{D})+(1-\delta)\mathcal{L}_{\text{KD}}(\mathcal{D})$
(\S~\ref{sec:method}).
We use $\delta=1.0$ when KD is not enabled. Optionally, when KD is enabled, we set $\delta=0.7$, and KD temperature $T=3.0$.
The teacher is a CTC model trained on word+tone targets using the same XLS-R backbone,\footnote{\url{https://github.com/facebookresearch/fairseq/blob/main/examples/wav2vec/xlsr/README.md}} following the standard ASR adaptation procedure for XLS-R. Decoding uses lexicon constraint.

\paragraph{Optimization.}
We optimize all models with Adam (learning rate $5\times10^{-4}$), weight decay $0.01$, and gradient clipping at $1.0$.
We use a batch size of 4 with 4-step gradient accumulation (effective batch size 16), and train for 12{,}000 update steps with 1{,}200 warmup steps.
All experiments are run on a single NVIDIA L40 GPU (48\,GB).

\subsection{Layer indexing and freezing convention}
\label{app:layer_indexing_freezing}

XLS-R consists of $M$ Transformer blocks (we use $M{=}24$), indexed from bottom to top as $\{1,\dots,M\}$.
Let $h^{(\ell)}$ denote the hidden states \emph{after} block $\ell$.
We extract token embeddings by pooling $h^{(\ell)}$ for all representation evaluations (retrieval, tone geometry, and tone classification).
Thus, ``19-layer'' and ``21-layer'' refer \emph{only} to the extraction block index $\ell$ (not the number of updated blocks).

We consider two training modes.
In the \textbf{freeze} setting, we freeze the bottom 12 blocks throughout training and never update them.
In the \textbf{non-freeze} setting, these blocks are also trainable.

Our two-stage pipeline allocates the remaining capacity around the extraction block $\ell$.
Stage~1 updates blocks 13--$\ell$ to learn speaker-invariant and tone-aware geometry at the representation space.
Stage~2 fine-tunes the upper blocks $(\ell{+}1)$--$M$ and CTC head for ASR, while keeping blocks 1--$\ell$ fixed
so that the Stage~1 representation interface remains unchanged during ASR adaptation.

\subsection{Evaluation Metrics}
\label{app:metrics}

\noindent \textbf{Embedding similarity.}
Given an utterance $x$, we extract a pooled normalized embedding
$\vz=f_\theta(x)$.
We use cosine similarity
\[
\mathrm{sim}(\vu,\vv)=\vu^\top \vv ,
\]
where $u,v$ are $\ell_2$-normalized vectors.

\noindent \textbf{Tone-geometry statistics.}
We report average cosine similarity for \emph{positives} (same lexical item, same tone),
\[
\mathrm{PosSim}
=
\mathbb{E}_{(i,j)\in \mathcal{A}_{\text{pos}}}\!\left[\mathrm{cos}(\vz_i,\vz_j)\right],
\]
and average cosine distance for negatives,
\begin{align*}
d_{\cos}(\vu,\vv) &= 1-\mathrm{cos}(\vu,\vv), \\
\mathrm{NegDist}
&=
\mathbb{E}_{(i,j)\in \mathcal{A}_{\text{neg}}}\!\left[d_{\cos}(\vz_i,\vz_j)\right].
\end{align*}
We instantiate $\mathcal{A}_{\text{neg}}$ as (i) \emph{hard} negatives: same lexical item but different tone, and
(ii) \emph{soft} negatives: different lexical items (therefore often also different tones).
Since $\mathrm{cos}(u,v)\in[-1,1]$ for normalized vectors, $d_{\cos}(u,v)\in[0,2]$.

\noindent \textbf{Cross-gender retrieval accuracy.}
For each query embedding $z_q$, we rank all opposite-gender gallery embeddings by $\mathrm{sim}(z_q,z_g)$.
Top-$K$ accuracy is the fraction of queries for which at least one of the $K$ highest-ranked gallery items
matches the query's lexical label.

\noindent \textbf{ASR error rates.}
We report word error rate (WER) and character error rate (CER), computed as normalized Levenshtein distance:
\begin{align*}
\mathrm{WER} &= \frac{S_w + D_w + I_w}{N_w}, \\
\mathrm{CER} &= \frac{S_c + D_c + I_c}{N_c},
\end{align*}
where $S$, $D$, and $I$ are the numbers of substitutions, deletions, and insertions, and $N$ is the number
of reference tokens (words for WER, characters for CER).

\subsection{Embedding Extraction and Pooling}
\label{app:pooling}

Across all experiments, we extract frame-level hidden states from the adapted encoder's block $\ell$ for each word token segment, and then pool over time to obtain a fixed-dimensional token embedding.
Let $\{h_t\}_{t=1}^{T} \subset \mathbb{R}^{d}$ denote the frame-level embeddings aligned to a token segment.
We consider the following pooling operators:
\begin{flalign*}
&\text{mean:}\quad \vz = \frac{1}{T}\sum_{t=1}^{T} \vh_t,\\
&\text{max:}\quad \vz_k = \max_{t \in \{1,\dots,T\}} (\vh_t)_k,\ \forall k \in \{1,\dots,d\},\\
&\text{max+mean:}\quad \vz = \frac{\vz^{\text{max}}+ \vz^{\text{mean}}}{2},\\
&\text{weighted:}\quad \vz = \alpha\, \vz^{\text{max}} + (1-\alpha)\, \vz^{\text{mean}},\ \alpha=0.7.
&&
\end{flalign*}

\paragraph{Default pooling choices.}
We use \textbf{max pooling} for \textbf{cross-gender retrieval} throughout the paper, as it yields the most stable rankings for lexical identity under speaker variation. For all \textbf{tone-related analyses} (tone classification and tone geometry) and all \textbf{similarity-scoring experiments}, we use \textbf{mean pooling}, which better reflects the overall segment-level representation and produces smoother similarity estimates. These pooling choices are \emph{fixed globally} for \textbf{SITA} and are not tuned per experiment, per model variant, or per dataset.

\paragraph{Pooling ablation.}
For completeness, we provide an ablation over pooling strategies in Table~\ref{tab:hmong_pooling_ablation}, confirming that while retrieval performance varies with the pooling operator, our main qualitative conclusions remain unchanged.

\subsection{Similarity scorer protocol and pitch-shift setup}
\label{app:sim_protocol}

We evaluate whether token embeddings behave sensibly as a similarity scorer by measuring cosine similarity
$s(\vz_a,\vz_b) = \frac{\vz_a^\top \vz_b}{\|\vz_a\|\|\vz_b\|}$
under controlled pairing protocols.
Unless noted otherwise, all similarity-scoring experiments use \textbf{mean-pooled} token embeddings (Appendix~\ref{app:pooling}).

\paragraph{Experiment 1: correct-word similarity (cross-speaker).}
We form pairs of the \emph{same base word and same tone} spoken by \emph{different speakers} and report the mean cosine similarity. This approximates reference-based pronunciation matching: similarity should be high for correct productions (same word and tone) across speakers.

\paragraph{Experiment 2: pitch-shift robustness.}
To test robustness to purely acoustic perturbations that preserve lexical identity, we apply synthetic pitch shifts to the waveform and recompute embeddings.
We use semitone shifts in $\{-2, -1, -0.5, +0.5, +1, +2\}$ and report the mean cosine similarity between the original token embedding and each pitch-shifted version.

\paragraph{Experiment 3: wrong-word similarity.}
We sample pairs of \emph{different base words} via permutation and report the mean cosine similarity. A good scorer should assign low similarity to such randomly mismatched lexical content.

\paragraph{Experiment 4: tone-confusion similarity.}
We sample pairs of the \emph{same base word but different tones} and report the mean cosine similarity. This directly probes whether the embedding space collapses tonal contrasts, and a tone-aware representation should reduce similarity relative to Exp1.

\begin{figure}[t]
\centering
\includegraphics[width=\linewidth]{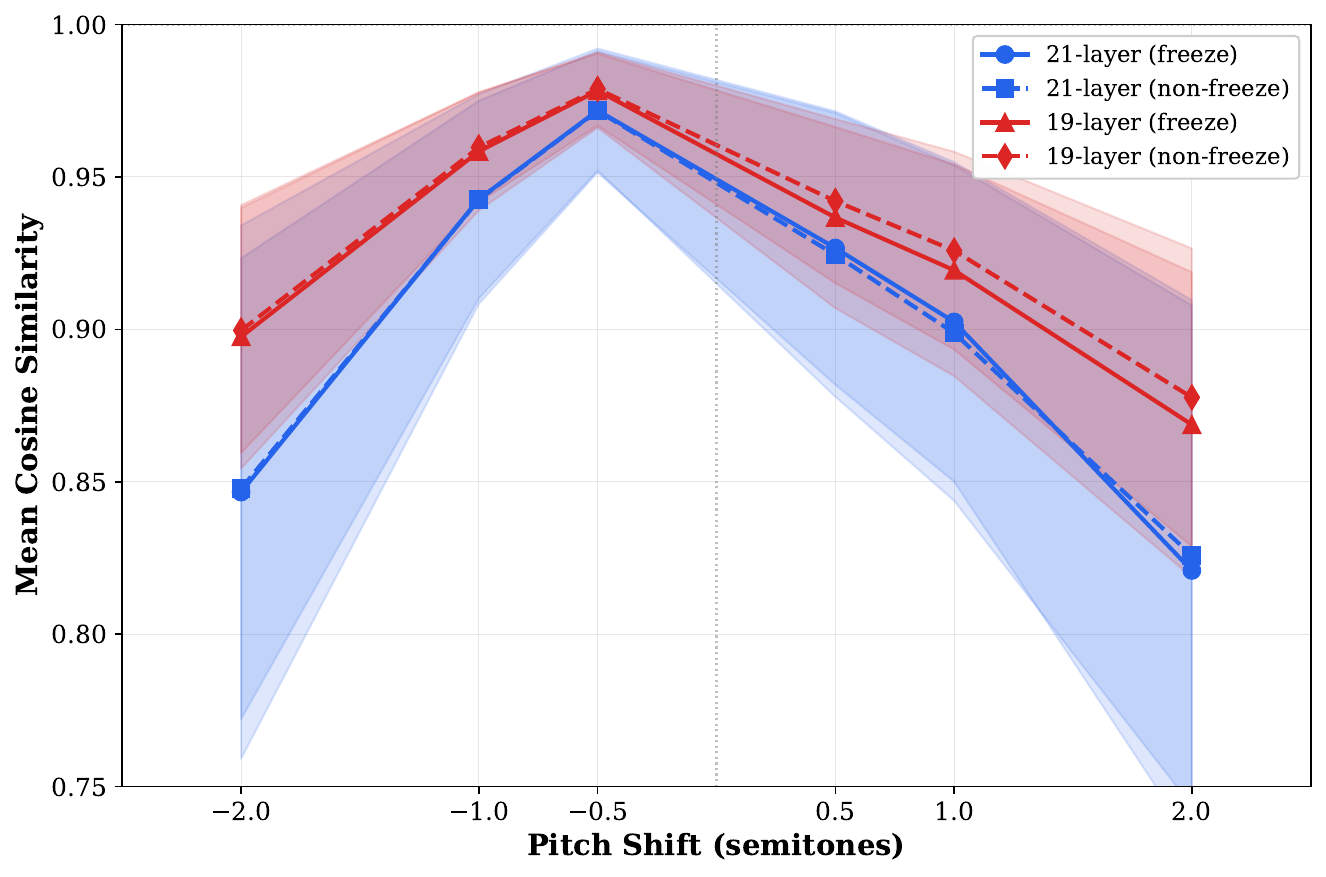}
\caption{
Tone perturbation robustness on Hmong word embeddings.
For each model, we report the mean cosine similarity between the original word and versions with pitch shifts of $-2$, $-1$, $-0.5$, $+0.5$, $+1$, and $+2$ semitones.
All four variants maintain high similarity under mild shifts (e.g., $\pm 0.5$), indicating robustness to small acoustic fluctuations, while similarity drops as the pitch shift magnitude increases, especially for the larger $\pm 2$ semitone changes.
The 21-layer models show a slightly steeper decay than the 19-layer models, suggesting stronger sensitivity to large tone perturbations while remaining stable under realistic within-speaker variation.
}
\label{fig:hmong_tone_perturbation}
\end{figure}

\begin{table*}[t]
\centering
\small
\caption{Effect of pooling strategy on cross-gender word retrieval on Hmong. We report Top-1 / Top-5 accuracy for querying female speech against a male gallery (F$\rightarrow$M) and male speech against a female gallery (M$\rightarrow$F) under different embedding pooling choices.}
\label{tab:hmong_pooling_ablation}
\begin{tabular}{llcccc}
\toprule
Encoder Depth & Pooling & F$\rightarrow$M Top-1 & F$\rightarrow$M Top-5 & M$\rightarrow$F Top-1 & M$\rightarrow$F Top-5 \\
\midrule
\multicolumn{6}{c}{\textbf{21-layer, lower layers frozen}} \\
\midrule
21 (freeze) & mean                & 0.5714 & 0.9000 & 0.5000 & 0.8071 \\
21 (freeze) & max                 & 0.6286 & 0.9286 & 0.5929 & \textbf{0.8893} \\
21 (freeze) & max+mean            & 0.6143 & 0.9429 & 0.5500 & 0.8536 \\
21 (freeze) & max+mean (weighted) & 0.6143 & 0.9429 & 0.5643 & 0.8643 \\
\midrule
\multicolumn{6}{c}{\textbf{21-layer, no freezing}} \\
\midrule
21 (non-freeze) & mean                & 0.6429 & 0.9000 & 0.5500 & 0.8429 \\
21 (non-freeze) & max                 & \textbf{0.6714} & \textbf{0.9571} & \textbf{0.6250} & 0.8821 \\
21 (non-freeze) & max+mean            & 0.6857 & 0.9143 & 0.5964 & 0.8714 \\
21 (non-freeze) & max+mean (weighted) & 0.6714 & 0.9286 & 0.6071 & 0.8786 \\
\midrule
\multicolumn{6}{c}{\textbf{19-layer, lower layers frozen}} \\
\midrule
19 (freeze) & mean                & 0.4429 & 0.9143 & 0.4250 & 0.7536 \\
19 (freeze) & max                 & 0.5286 & 0.9429 & 0.5643 & 0.8571 \\
19 (freeze) & max+mean            & 0.4857 & 0.9286 & 0.5214 & 0.8036 \\
19 (freeze) & max+mean (weighted) & 0.4857 & 0.9286 & 0.5393 & 0.8286 \\
\midrule
\multicolumn{6}{c}{\textbf{19-layer, no freezing}} \\
\midrule
19 (non-freeze) & mean                & 0.5143 & 0.8857 & 0.4000 & 0.7429 \\
19 (non-freeze) & max                 & 0.5143 & 0.9000 & 0.5286 & 0.8571 \\
19 (non-freeze) & max+mean            & 0.5286 & 0.9000 & 0.4929 & 0.8214 \\
19 (non-freeze) & max+mean (weighted) & 0.5143 & 0.9000 & 0.5250 & 0.8464 \\
\bottomrule
\end{tabular}
\end{table*}

\section{Additional Analyses on Hmong}


\begin{figure}[t]
    \centering
    \includegraphics[width=\linewidth]{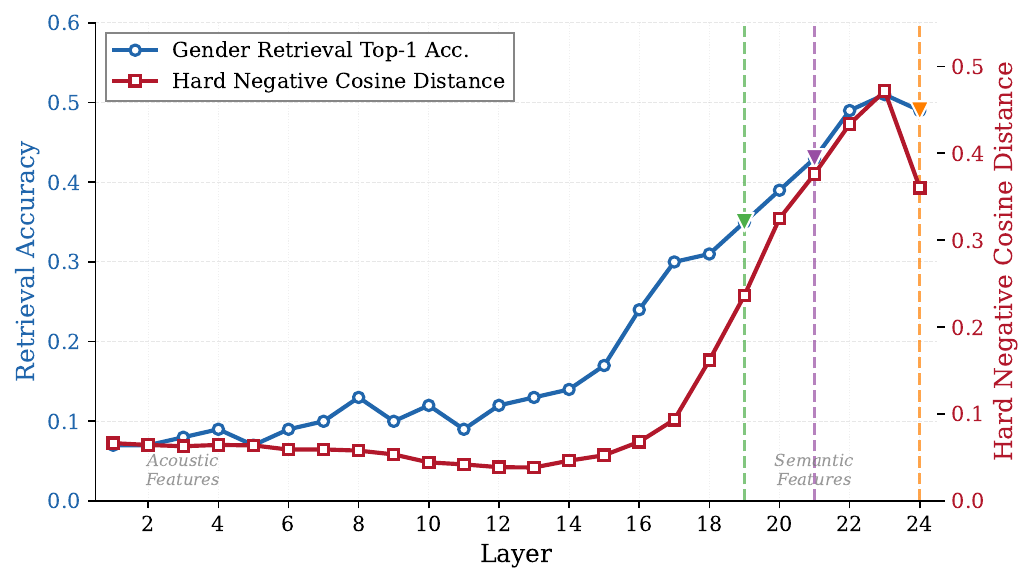}
    \caption{Layer-wise probing on a Stage 1 SITA checkpoint trained with layer 24 as the embedding extraction layer. Left axis: Avg Top-1 cross-gender retrieval. Right axis: hard negative cosine distance. Vertical markers indicate the selected embedding extraction layers, which balance strong retrieval/tone separation with sufficient upper-layer capacity for Stage~2 ASR training.}
    \label{fig:layer_analysis}
\end{figure}

\subsection{Embedding Extraction Layer Selection}
\label{app:feature_layer_selection}
Figure ~\ref{fig:layer_analysis} shows probing for every encoder layer on a SITA checkpoint trained with embedding extraction layer 24 by extracting pooled embeddings from layer $\ell\in\{1,\dots,24\}$ and measuring Avg Top-1 cross-gender retrieval and hard-negative cosine distance. Deeper layers improve both metrics substantially. However, choosing the very last layers as the embedding extraction layer leaves little capacity for Stage~2 ASR updates without directly overwriting the evaluated embedding space. We therefore select layers 19 and 21 as practical trade-offs: layer 19 is where both metrics begin to rise sharply while still leaving five upper layers for Stage 2, and layer 21 further improves retrieval/tone separation.

\subsection{Ablation Study on Voice Conversion Augmentation}
\label{app:freevc_ablation}

\paragraph{FreeVC quality inspection.}
Since our converted clips are short isolated word recordings, FreeVC operates in a favorable regime compared to long-form spontaneous speech. To verify that conversion artifacts are not driving the gains, we ran an audit on all converted clips using three measures. Signal-to-noise ratio of the converted audio. Mel-spectrogram cosine similarity between the original and converted clip. Duration ratio of the converted clip over the original. The audit shows high-fidelity conversion with minimal temporal distortion. Mean SNR is $13.91$ dB with standard deviation $4.18$, minimum $3.54$ and maximum $32.29$. Mean mel cosine similarity is $0.9259$ with standard deviation $0.0291$, minimum $0.7597$ and maximum $0.9879$. Mean duration ratio is $0.9968$ with standard deviation $0.0106$, and $90\%$ of pairs match the original duration exactly. We did not apply additional quality filtering after the audit.

To test whether our gains rely on voice conversion, we retrain our main frozen setting (21-layer, freeze) \emph{without} FreeVC, keeping all other training settings and waveform perturbations unchanged.
Table~\ref{tab:ablation_freevc} removes FreeVC and uses only on-the-fly waveform perturbations.
Removing FreeVC does not reduce performance and can even improve cross-gender retrieval in one direction, while tone-geometry statistics remain comparable (Hard/Soft NegDis stay high).
This suggests that FreeVC is an optional augmentation rather than a requirement for SITA: in this low-resource, speaker-imbalanced setting, synthetic voice conversion may introduce distributional artifacts that help some query--gallery compositions but are not consistently beneficial across directions.
Importantly, our main qualitative conclusions, namely strong cross-gender lexical retrieval and reduced tone collapse, do not depend only on voice conversion.

\begin{table*}[h]
\centering
\small
\setlength{\tabcolsep}{4.5pt}
\caption{Effect of removing FreeVC on the main Hmong setting (21-layer, frozen). ``w/o FreeVC'' uses only waveform perturbations. Higher is better for Top-$K$ and NegDis; PosSim is cosine similarity.}
\label{tab:ablation_freevc}
\begin{tabular}{lcccccc}
\toprule
Setting & F$\rightarrow$M Top-1 & F$\rightarrow$M Top-5 & M$\rightarrow$F Top-1 & M$\rightarrow$F Top-5 & PosSim & Hard/Soft NegDis \\
\midrule
w/ FreeVC & 0.6286 & 0.9286 & 0.5929 & 0.8893 & 0.8000 & 0.6748/0.9398 \\
w/o FreeVC & 0.6143 & 0.9000 & 0.7988 & 0.9825 & 0.7390 & 0.6816/0.9520 \\
\bottomrule
\end{tabular}
\end{table*}

\subsection{Pooling Strategy Ablation}
\label{app:pooling_ablation}

\textbf{SITA} operates on frame-level encoder outputs $\{h_t\}_{t=1}^{T}$ and requires a pooling operator to obtain a fixed-dimensional token embedding $z$ (Appendix~\ref{app:pooling}).
We additionally compare mean pooling, concatenated max+mean pooling, and a weighted max+mean. Table~\ref{tab:hmong_pooling_ablation} reports Top-1/Top-5 retrieval accuracy under different pooling choices across adaptation depths and freezing settings.
Overall, \textbf{max pooling consistently yields the strongest retrieval performance} for both F$\rightarrow$M and M$\rightarrow$F, especially in the lightweight frozen setting, where 21-layer freeze improves from 0.57/0.50 Top-1 under mean to 0.63/0.59 under max.
Concatenating max+mean and using a weighted mixture ($\alpha=0.7$ for max) provide competitive results but do not consistently outperform pure max pooling.
In contrast, mean pooling tends to underperform for retrieval, likely because averaging smooths out salient phonetic cues that are critical for distinguishing short word tokens across speakers. Based on this ablation, we adopt \textbf{max pooling for retrieval} throughout the paper.
For tone geometry and similarity-scoring analyses, we use \textbf{mean pooling} to obtain smoother, more stable embeddings under small acoustic perturbations (Appendix~A.2), unless otherwise stated.

\subsection{Margin-Based Tone Repulsion Loss}
\label{app:tone_margin}

In addition to the main InfoNCE-based tone objective, we experimented with a margin-based formulation that explicitly enforces separation between tones via hinge penalties. Let $\vz_i\in\mathbb{R}^d$ denote the normalized embedding for token $i$ and $s_{ij}=\mathrm{sim}(\vz_i,\vz_j)=\vz_i^\top \vz_j$. For each anchor $i$, we form three pair sets: positives $\gP(i)$ (same word, same tone), hard negatives $\gH(i)$ (same word, different tone), and soft negatives $\gS(i)$ (different word). The per-anchor loss is
\begin{equation}
\label{eq:tone_margin_anchor}
\begin{aligned}
\ell_i^{\text{margin}}
&=
\lambda_{\text{attr}}
\cdot
\frac{1}{|\gP(i)|}
\sum_{j\in\gP(i)}
\bigl(-s_{ij}\bigr)
\\
&\quad+
\lambda_{\text{hard}}
\cdot
\frac{1}{|\gH(i)|}
\sum_{j\in\gH(i)}
\bigl[s_{ij}-m_{\text{hard}}\bigr]_+
\\
&\quad+
\lambda_{\text{soft}}
\cdot
\frac{1}{|\gS(i)|}
\sum_{j\in\gS(i)}
\bigl[s_{ij}-m_{\text{soft}}\bigr]_+ ,
\end{aligned}
\end{equation}
where $[x]_+=\max(x,0)$. We set $m_{\text{hard}}<m_{\text{soft}}$ to impose a stricter separation requirement for same-word different-tone pairs than for different-word pairs. The full objective averages over anchors:
\begin{equation}
\label{eq:tone_margin}
\mathcal{L}_{\text{tone}}^{\text{margin}}
=
\frac{1}{|\mathcal{I}|}
\sum_{i\in\mathcal{I}}
\ell_i^{\text{margin}} .
\end{equation}

\begin{table}[t]
\centering
\small
\setlength{\tabcolsep}{3.5pt}
\renewcommand{\arraystretch}{0.95}
\caption{Margin-based tone loss ablation on Hmong (\S\ref{app:tone_margin}). Retrieval: Avg Top-1/Top-5. Tone geometry: cosine distance on hard/soft negatives.}
\label{tab:ablation_margin}
\begin{tabular}{lccccc}
\toprule
Loss & Emb. & Avg T1 & Avg T5 & Hard & Soft \\
\midrule
InfoNCE & 21 & 0.6108 & 0.9088 & 0.6748 & 0.9398 \\
Margin  & 21 & 0.8728 & 0.9411 & 0.9523 & 1.0115 \\
\midrule
InfoNCE & 19 & 0.5465 & 0.9000 & 0.4662 & 0.6856 \\
Margin  & 19 & 0.8571 & 0.9484 & 0.6456 & 0.8601 \\
\bottomrule
\end{tabular}
\end{table}

This margin loss directly targets \emph{tone collapse} by pushing hard negatives below a similarity threshold while preserving within-tone cohesion through the attractive term. 

In our margin-based experiment, we set
$m_{\text{hard}}=-0.1$, $m_{\text{soft}}=0.1$,
and weights $\lambda_{\text{attr}}=1.0$,
$\lambda_{\text{hard}}=0.5$,
$\lambda_{\text{soft}}=1.0$.

Table~\ref{tab:ablation_margin} shows that the margin-based loss produces stronger tone repulsion than InfoNCE under the same adaptation setting: it increases cosine distance for both hard and soft negatives. For our embedding extraction layer 21 setting, it increases hard cosine distance from  0.675 to 0.952, soft from 0.940 to 1.012, and correspondingly improves in-domain retrieval. However, these gains come with worse robustness under speaker shift (cf.\ \S\ref{app:unseen_full}), so we keep InfoNCE as the default and report Margin only as an auxiliary ablation.

\subsection{Tone Classification Results}
\label{app:tone_cls}

In \textbf{SITA}, we include a tone-classification head during Stage 1 training, weighted by $\lambda$ as in \S\ref{sec:method}. In this appendix, we report its Top-1 and Top-3 accuracy for all our settings. We use these results as a sanity check that the adapted embeddings retain decodable tone cues, but we emphasize tone-geometry metrics in the main paper since they better capture \emph{relative} separation between same-word different-tone pairs. Table~\ref{tab:app_tone_cls_seen} shows that tone is consistently predictable from our embeddings across adaptation depths and freezing policies.

\begin{table}[t]
\centering
\small
\caption{Tone classification on Hmong. We report Top-1/Top-3 accuracy
of the auxiliary tone head.}
\label{tab:app_tone_cls_seen}
\begin{tabular}{lcc}
\toprule
Setting & Top-1 (\%) & Top-3 (\%) \\
\midrule
Tone Cls Only & 86.00 & 96.29 \\
SITA 21-no-freeze & 85.14 & 96.57 \\
SITA 21-freeze & 85.43 & 97.71 \\
SITA 19-no-freeze & 83.43 & 96.57 \\
SITA 19-freeze & 84.57 & 97.71 \\
\bottomrule
\end{tabular}
\end{table}



\subsection{Similarity Score Evaluation}
\label{app:sim_eval}

Beyond aggregate retrieval and tone metrics, we evaluate whether \textbf{SITA} embeddings behave sensibly as a similarity scorer.
Given a reference pronunciation of a target word, a useful embedding space should assign high similarity to correct repetitions by different speakers, but lower similarity to (i) different tones of the same base word and (ii) completely different words. See Appendix~\ref{app:sim_protocol} for the experiment setup.

\begin{table}[t]
\centering
\small
\setlength{\tabcolsep}{4.2pt}
\renewcommand{\arraystretch}{0.95}
\caption{Similarity scoring on Hmong embeddings. E1: same word+tone (higher better). E3: different word (lower). E4: same word, different tone (lower).}
\label{tab:hmong_sim}
\begin{tabular}{lccc}
\toprule
Model & E1 $\uparrow$ & E3 $\downarrow$ & E4 $\downarrow$ \\
\midrule
SITA-21 (freeze)     & 0.7993 & \textbf{0.1045} & \textbf{0.3272} \\
SITA-21 (unfreeze)   & 0.7955 & 0.1046 & 0.3452 \\
SITA-19 (freeze)     & \textbf{0.8548} & 0.3442 & 0.5292 \\
SITA-19 (unfreeze)   & 0.8534 & 0.3624 & 0.5486 \\
\bottomrule
\end{tabular}
\end{table}

Table~\ref{tab:hmong_sim} shows that our 21-layer models preserve high similarity for correct repetitions while strongly suppressing similarity for wrong words and wrong tones. In contrast, the 19-layer variants yield much higher similarity for wrong words and wrong tones, indicating weaker discrimination. Finally, we evaluate robustness to purely acoustic perturbations (Experiment 2) by applying synthetic pitch shifts. As can be seen from Fig. \ref{fig:hmong_tone_perturbation}, similarity remains stable under mild shifts (e.g., $\pm0.5$--$1$ semitone) and decreases gradually for larger shifts, remaining above the ``wrong word'' baseline, suggesting robustness to benign prosodic variation while retaining sensitivity to tone errors.

\subsection{Per-Tone Analysis}
\label{app:per_tone}

We provide a per-tone diagnostic of \textbf{SITA} (21-freeze, InfoNCE-based tone objective) to identify which tones benefit most and which remain challenging. We report (i) per-tone tone classification accuracy and (ii) per-tone embedding geometry. For geometry, we compute \emph{PosSim} (cosine similarity for same-word same-tone pairs, $\uparrow$), \emph{Hard NegDist} (cosine distance for same-word different-tone pairs, $\uparrow$),
and \emph{Soft NegDist} (cosine distance for different-word pairs, $\uparrow$), reported as mean$\pm$std.

The largest errors concentrate on Tones 2 and 3 (Table~\ref{tab:per_tone_acc}).
Consistent with this, these tones show lower PosSim and larger variance (Table~\ref{tab:per_tone_geom}),
suggesting reduced within-tone consistency.
Hard NegDist varies less across tones, indicating that the main remaining challenge is tone-specific cohesion rather than global tone separation.

\begin{table}[t]
\centering
\small
\setlength{\tabcolsep}{4.0pt}
\renewcommand{\arraystretch}{0.95}
\caption{Per-tone tone classification accuracy on Hmong (21-freeze, InfoNCE).}
\label{tab:per_tone_acc}
\begin{tabular}{lc}
\toprule
Tone & Accuracy \\
\midrule
1 (b) & 0.98 \\
2     & 0.74 \\
3 (s) & 0.66 \\
4 (j) & 0.90 \\
5 (v) & 0.94 \\
6 (g) & 0.86 \\
7 (m) & 0.90 \\
\bottomrule
\end{tabular}
\end{table}


\begin{table}[t]
\centering
\small
\setlength{\tabcolsep}{3.5pt}
\renewcommand{\arraystretch}{0.95}
\caption{Per-tone embedding geometry on Hmong (21-freeze, InfoNCE). PosSim: cosine similarity for positives ($\uparrow$). Hard/Soft NegDist: cosine distance for hard/soft negatives ($\uparrow$). We report mean$\pm$std.}
\label{tab:per_tone_geom}
\begin{tabular}{lccc}
\toprule
Tone & PosSim $\uparrow$ & Hard NegDist $\uparrow$ & Soft NegDist $\uparrow$ \\
\midrule
1 (b) & 0.84$\pm$0.046 & 0.71$\pm$0.074 & 0.90$\pm$0.016 \\
2     & 0.74$\pm$0.070 & 0.63$\pm$0.068 & 0.90$\pm$0.014 \\
3 (s) & 0.77$\pm$0.062 & 0.63$\pm$0.069 & 0.89$\pm$0.013 \\
4 (j) & 0.80$\pm$0.048 & 0.68$\pm$0.074 & 0.89$\pm$0.011 \\
5 (v) & 0.82$\pm$0.050 & 0.66$\pm$0.065 & 0.88$\pm$0.017 \\
6 (g) & 0.74$\pm$0.072 & 0.69$\pm$0.094 & 0.90$\pm$0.015 \\
7 (m) & 0.80$\pm$0.062 & 0.69$\pm$0.092 & 0.89$\pm$0.012 \\
\bottomrule
\end{tabular}
\end{table}

\subsection{ASR Case Study on the Base Word \texttt{lia}}
\label{app:case_study}

Table~\ref{tab:case_study} accompanies the qualitative analysis in \S\ref{sec:results_tone}. We decode all seven tone variants of the base word \texttt{lia} and compare an off-the-shelf baseline with \method. Whisper decodes with an open vocabulary and its outputs are shown verbatim. Whisper never recovers the base word and instead falls back on English proper nouns, which is the decoding-level signature of the tone collapse visible in Figure~\ref{fig:case_study}. \method stays within the correct base word on every token and its remaining errors are tone confusions, most of them within the falling tone group.

\begin{table}[h]
\centering
\setlength{\tabcolsep}{4.0pt}
\renewcommand{\arraystretch}{0.95}
\caption{
ASR case study on Hmong tone-marked targets for base word \texttt{lia}.
We compare an off-the-shelf baseline, Whisper-large-v3 with open-vocabulary decoding, against \method.
Whisper outputs are shown verbatim.
}
\label{tab:case_study}
\resizebox{0.48\textwidth}{!}{
\begin{tabular}{c l l l}
\toprule
\textbf{Tone} & \textbf{Ref. (word+tone)} & \textbf{Whisper-large-v3} & \textbf{SITA} \\
\midrule
1 (b) & \texttt{liab} & \texttt{Leah?} & \texttt{liab} \\
2     & \texttt{lia}  & \texttt{Leah}  & \texttt{liam} \\
3 (s) & \texttt{lias} & \texttt{LEARN!} & \texttt{lias} \\
4 (j) & \texttt{liaj} & \texttt{Leah}  & \texttt{liav} \\
5 (v) & \texttt{liav} & \texttt{Leah}  & \texttt{liav} \\
6 (g) & \texttt{liag} & \texttt{moves} & \texttt{liav} \\
7 (m) & \texttt{liam} & \texttt{Leia.} & \texttt{liag} \\
\bottomrule
\end{tabular}}
\end{table}

\section{Additional Experiments and Full Result Tables}
\label{app:full_tables}

This appendix reports additional \textbf{SITA} results omitted from the main paper for clarity:
(i) cross-backbone transfer to WavLM,
(ii) a non-tonal English regression probe,
(iii) a sentence-level probe on pseudo-sentences,
(iv) full retrieval and tone-geometry tables including non-frozen variants, InfoNCE and margin variants, and optional knowledge distillation for ASR, and
(v) full tables for unseen-speaker evaluation.

\subsection{Cross-Backbone Transfer Experiment: SITA on WavLM}
\label{app:wavlm_backbone}

We test whether the \method recipe transfers across SSL backbones by replacing the XLS-R encoder with WavLM Large \cite{Chen_2022} and keeping the same objectives and training schedule. We use the same Hmong corpus, the same gender, tone, and word labels, and the same frozen lower-block setup. The goal is not to claim a new state of the art, but to check that the recipe is not specific to one backbone.

\begin{table*}[h]
\centering
\small
\setlength{\tabcolsep}{6pt}
\renewcommand{\arraystretch}{0.95}
\caption{Cross-backbone transfer of the \method recipe. Same objectives and same training schedule, only the SSL backbone is changed. Retrieval is cross-gender word retrieval on Hmong. Hard-neg cosine distance measures tone separation for same word with different tones.}
\label{tab:wavlm_backbone}
\begin{tabular}{lccccc}
\toprule
\textbf{Backbone} & \textbf{F$\rightarrow$M T1} & \textbf{F$\rightarrow$M T5} & \textbf{M$\rightarrow$F T1} & \textbf{M$\rightarrow$F T5} & \textbf{Hard-neg dist.} \\
\midrule
SITA + XLS-R  & 0.6286 & 0.9286 & 0.5929 & 0.8893 & 0.6748 \\
SITA + WavLM  & 0.5143 & 0.8286 & 0.5107 & 0.8214 & 0.3379 \\
\bottomrule
\end{tabular}
\end{table*}

Table~\ref{tab:wavlm_backbone} reports the comparison. \method improves both backbones, but XLS-R remains the stronger base in our setting. We attribute the gap to differences in the pretraining priors. XLS-R was pretrained on 128 languages and tends to be a better starting point when the target language is underrepresented or absent from English-centric corpora. WavLM was pretrained mostly on English speech. The takeaway is that \method is compatible with multiple SSL backbones, but the pretrained checkpoint and its embedding geometry under our word-level cross-speaker setting matter for final performance.

\subsection{Non-Tonal Language Probe: Speech Commands}
\label{app:non_tonal_probe}

To check whether \method, trained with Hmong tone supervision, degrades the encoder on a non-tonal language, we evaluate it on English Speech Commands v0.02~\cite{warden2018speechcommandsdatasetlimitedvocabulary}. From the test split, we randomly sample 20 keywords and up to 10 speakers per keyword with one clip per speaker, giving 198 clips in total (fixed random seed). For both SITA Stage 1 and vanilla XLS-R 300M, we extract frozen embeddings at layer 21 with mean pooling. For each query clip we rank all other clips by cosine similarity, restricted to clips from a different speaker. A query is correct at Top-$K$ if any of the top-$K$ ranked clips has the same keyword label. Reported numbers are averaged over all 198 queries.

\begin{table}[h]
\centering
\small
\setlength{\tabcolsep}{6pt}
\renewcommand{\arraystretch}{0.95}
\caption{Non-tonal probe on English Speech Commands. Cross-speaker word retrieval Top-1 and Top-5 with frozen embeddings at layer 21.}
\label{tab:non_tonal_probe}
\begin{tabular}{lcc}
\toprule
\textbf{Model} & \textbf{Top-1} & \textbf{Top-5} \\
\midrule
Vanilla XLS-R   & 0.3485 & 0.7273 \\
\textbf{SITA Stage 1} & \textbf{0.6263} & \textbf{0.8586} \\
\midrule
Delta (SITA $-$ XLS-R) & $+0.2778$ & $+0.1313$ \\
\bottomrule
\end{tabular}
\end{table}

As Table~\ref{tab:non_tonal_probe} shows, \method does not regress on this non-tonal benchmark. It in fact improves cross-speaker word retrieval over vanilla XLS-R by a clear margin. We attribute this to the cross-speaker contrastive objective being language-agnostic, since pulling same-word cross-speaker pairs together while pushing different lexical items apart is useful for keyword spotting in any language. We treat this as a small-scale sanity probe rather than a full non-tonal benchmark. A broader study covering longer-form non-tonal speech and a wider set of downstream tasks is left as future work.

\subsection{Sentence-Level Probe via Pseudo-Sentences}
\label{app:sentence_probe}

Our main evaluation uses isolated word recordings. To check whether \method's tone awareness and speaker invariance hold in a longer audio context, we construct controlled pseudo-sentences by concatenating Hmong word clips and test how each model reacts to a meaning-changing tone swap versus a meaning-preserving speaker swap.

\paragraph{Trial construction.}
For each trial we pick a target base word that has at least two tone variants in the Hmong training metadata, then pick three other base words as context. The target sits at position 2 in a 4-word sequence. We build three pseudo-sentences that differ only in the target slot:
\begin{itemize}[leftmargin=13pt,itemsep=1pt,topsep=2pt]
  \item $S_\text{original}$: target word with tone $i$, spoken by speaker $A$.
  \item $S_\text{tone}$: same base word with tone $j \neq i$, any speaker. In a tonal language this changes the meaning of the target.
  \item $S_\text{speaker}$: same target word with tone $i$, spoken by a different speaker $B \neq A$. The meaning is preserved.
\end{itemize}
Words are concatenated with 300\,ms of silence between them. We sample 200 trials with a fixed random seed.

\paragraph{Metric.}
For each model we extract pooled embeddings at layer 21 with mean pooling on the full pseudo-sentence. Let $\vz_o, \vz_t, \vz_s$ denote the pooled sentence embeddings of $S_\text{original}$, $S_\text{tone}$, and $S_\text{speaker}$. We define
\begin{align*}
d_\text{tone}    &= 1 - \cos(\vz_o,\, \vz_t), \\
d_\text{speaker} &= 1 - \cos(\vz_o,\, \vz_s).
\end{align*}
A model that is both tone-aware and speaker-invariant should produce $d_\text{tone} \gg d_\text{speaker}$ in this longer-context setting. We compare vanilla XLS-R 300M, LASR, UniSpeech, and SITA Stage 1, all used as frozen feature extractors.

\begin{table}[h]
\centering
\small
\setlength{\tabcolsep}{6pt}
\renewcommand{\arraystretch}{0.95}
\caption{Sentence-level probe on Hmong pseudo-sentences. Mean cosine distance over 200 trials. Higher $d_\text{tone}$, lower $d_\text{speaker}$, and higher ratio are better.}
\label{tab:sentence_probe}
\begin{tabular}{lccc}
\toprule
\textbf{Model} & $d_\text{tone}$ $\uparrow$ & $d_\text{speaker}$ $\downarrow$ & \textbf{Ratio} $\uparrow$ \\
\midrule
Vanilla XLS-R & 0.0046 & 0.0046 & 0.99 \\
LASR          & 0.0782 & 0.0795 & 0.98 \\
UniSpeech     & 0.0228 & 0.0242 & 0.94 \\
\textbf{SITA} & \textbf{0.1347} & 0.1019 & \textbf{1.32} \\
\bottomrule
\end{tabular}
\end{table}

\paragraph{Analysis.}
Table~\ref{tab:sentence_probe} reports the results. SITA is the only model with a ratio above one, meaning a meaning-changing tone swap moves the sentence embedding strictly more than a meaning-preserving speaker swap. Vanilla XLS-R essentially does not move for either, consistent with the tone collapse pattern we observe at the word level. LASR and UniSpeech do respond to swaps but treat tone changes and speaker changes about equally, which is a fundamental failure in a tonal language where tone is lexical. These results indicate that the desired properties of \method, tone awareness and speaker invariance, are not narrowly word-level and carry over to longer audio contexts.

\paragraph{Caveat.}
These pseudo-sentences are concatenated isolated words with brief silence. They do not capture natural coarticulation or prosody. The probe is a controlled longer-context check, not a full continuous-speech evaluation. A broader study on natural continuous speech is future work.

\subsection{Full Retrieval and Tone Geometry Tables (Including Unfreezing)}
\label{app:full_tables_subsec}

Table~\ref{tab:hmong_full_allinone} consolidates Hmong results across retrieval, tone geometry, and ASR.
We include InfoNCE vs.\ margin/hinge, frozen vs.\ non-frozen updates, and optional KD for ASR.

\begin{table*}[t]
\centering
\small
\setlength{\tabcolsep}{4.2pt}
\renewcommand{\arraystretch}{0.95}
\caption{Full Hmong results table. Retrieval: Top-1/Top-5 for F$\rightarrow$M and M$\rightarrow$F. Tone geometry: PosSim and cosine distance on hard/soft negatives (Hard/Soft NegDist). ASR: CER/WER (lower is better). \textsc{Emb.}\ is the embedding extraction layer and \textsc{Upd.}\ is the Stage 1 update mode, either frozen (frz) or fully unfrozen (unfrz). KD applies to Stage 2 ASR training only (``--'' means not applicable).}
\label{tab:hmong_full_allinone}
\begin{tabular}{lcc|cccc|ccc|cc}
\toprule
\multirow{2}{*}{Emb.} & \multirow{2}{*}{Upd.} & \multirow{2}{*}{Loss}
& \multicolumn{4}{c|}{Retrieval}
& \multicolumn{3}{c|}{Tone Geometry}
& \multicolumn{2}{c}{ASR} \\
\cmidrule(lr){4-7}\cmidrule(lr){8-10}\cmidrule(lr){11-12}
& & 
& \multicolumn{2}{c}{F$\rightarrow$M} & \multicolumn{2}{c|}{M$\rightarrow$F}
& PosSim & Hard & Soft
& CER & WER \\
\cmidrule(lr){4-5}\cmidrule(lr){6-7}
& & & T1 & T5 & T1 & T5 & & & & & \\
\midrule
\multicolumn{12}{l}{\textbf{InfoNCE (main)}} \\
\midrule
21 & frz  & InfoNCE & 0.6286 & 0.9286 & 0.5929 & 0.8893 & 0.8000 & 0.6748 & 0.9398 & 0.2261 & 0.5505 \\
21 & frz+KD & InfoNCE & -- & -- & -- & -- & -- & -- & -- & 0.2301 & 0.5757 \\
21 & unfrz & InfoNCE & 0.6714 & 0.9571 & 0.6250 & 0.8821 & 0.7967 & 0.6566 & 0.9399 & 0.2152 & 0.5550 \\
21 & unfrz+KD & InfoNCE & -- & -- & -- & -- & -- & -- & -- & 0.2152 & 0.5436 \\
\midrule
19 & frz  & InfoNCE & 0.5286 & 0.9429 & 0.5643 & 0.8571 & 0.7997 & 0.4662 & 0.6856 & 0.2094 & 0.5344 \\
19 & frz+KD & InfoNCE & -- & -- & -- & -- & -- & -- & -- & 0.1985 & 0.5115 \\
19 & unfrz & InfoNCE & 0.5143 & 0.9000 & 0.5286 & 0.8571 & 0.7975 & 0.4531 & 0.6667 & 0.2140 & 0.5390 \\
19 & unfrz+KD & InfoNCE & -- & -- & -- & -- & -- & -- & -- & 0.2031 & 0.5183 \\
\midrule
\multicolumn{12}{l}{\textbf{Margin/Hinge (auxiliary)}} \\
\midrule
21 & frz  & Margin & 0.8214 & 0.9143 & 0.9242 & 0.9679 & 0.8649 & 0.9523 & 1.0115 & -- & -- \\
21 & unfrz & Margin & 0.6857 & 0.9286 & 0.7434 & 0.9738 & 0.6455 & 0.8213 & 0.9182 & -- & -- \\
19 & frz  & Margin & 0.7929 & 0.9143 & 0.9213 & 0.9825 & 0.8533 & 0.6456 & 0.8601 & -- & -- \\
19 & unfrz & Margin & 0.5500 & 0.8143 & 0.6560 & 0.9417 & 0.6398 & 0.4509 & 0.5661 & -- & -- \\
\bottomrule
\end{tabular}
\end{table*}

\begin{table*}[t]
\centering
\small
\setlength{\tabcolsep}{4.2pt}
\renewcommand{\arraystretch}{0.95}
\caption{Unseen-speaker split on Hmong (M$\rightarrow$F). Retrieval: Top-1/Top-5. Tone geometry: PosSim is cosine similarity on positives, and Hard/Soft are cosine distance on hard/soft negatives, where higher indicates stronger separation. \textsc{Emb.}\ is the embedding extraction layer and \textsc{Upd.}\ is frozen (frz) or unfrozen (unfrz) Stage 1 adaptation.}
\label{tab:unseen_full_all}
\begin{tabular}{lcc|cc|ccc}
\toprule
\multirow{2}{*}{Loss} & \multirow{2}{*}{Emb.} & \multirow{2}{*}{Upd.}
& \multicolumn{2}{c|}{Retrieval} & \multicolumn{3}{c}{Tone Geometry} \\
\cmidrule(lr){4-5}\cmidrule(lr){6-8}
& & & Top-1 & Top-5 & PosSim & Hard & Soft \\
\midrule
InfoNCE & 21 & unfrz & 0.5420 & 0.7809 & 0.7759 & 0.4245 & 0.7735 \\
InfoNCE & 21 & frz   & 0.5651 & 0.8089 & 0.7889 & 0.4406 & 0.7806 \\
InfoNCE & 19 & unfrz & 0.6145 & 0.8764 & 0.8588 & 0.2992 & 0.5314 \\
InfoNCE & 19 & frz   & \textbf{0.6870} & \textbf{0.9077} & 0.8653 & 0.3169 & 0.5554 \\
\midrule
Margin  & 21 & unfrz & 0.3888 & 0.6079 & 0.7966 & 0.5493 & 0.7496 \\
Margin  & 21 & frz   & 0.2965 & 0.4152 & 0.7954 & \textbf{0.5815} & \textbf{0.8442} \\
Margin  & 19 & unfrz & 0.4530 & 0.7759 & 0.8777 & 0.3010 & 0.4562 \\
Margin  & 19 & frz   & 0.4415 & 0.6837 & 0.8510 & 0.3484 & 0.6693 \\
\bottomrule
\end{tabular}
\end{table*}

\subsection{Unseen-Speaker Split: Full Retrieval and Tone Geometry}
\label{app:unseen_full}

Table~\ref{tab:unseen_full_all} summarizes the full unseen-speaker evaluation on Hmong (M$\rightarrow$F), reporting both cross-gender retrieval and tone-geometry statistics for frozen and non-frozen variants.
Overall, the standard InfoNCE objective remains most robust under speaker shift, while the margin-based objective increases tone separation with higher Hard/Soft cosine distance, but often reduces retrieval.

\section{Mandarin Supplementary Results}
\label{app:mandarin_full}

\begin{table}[t]
\centering
\small
\setlength{\tabcolsep}{3.6pt}
\renewcommand{\arraystretch}{0.95}
\caption{Mandarin: M$\rightarrow$F retrieval and tone geometry.}
\label{tab:zh_retr_tone_onecol}
\resizebox{\columnwidth}{!}{%
\begin{tabular}{lcc|cc|ccc}
\toprule
\multicolumn{3}{c|}{Setup} & \multicolumn{2}{c|}{Retrieval} & \multicolumn{3}{c}{Tone geom.} \\
\cmidrule(lr){1-3}\cmidrule(lr){4-5}\cmidrule(lr){6-8}
Setting & Feat & Frz & Top-1 & Top-5 & PosSim & Hard & Soft \\
\midrule
XLS-R              & -- & -- & 0.2457 & 0.4768 & 0.9856 & 0.0093 & 0.0194 \\
SSL Adaptation     & -- & -- & 0.0640 & 0.1780 & 0.9902 & 0.0056 & 0.0063 \\
Whisper (Large-v3) & -- & -- & 0.8476 & 0.9738 & 0.9876 & 0.0011 & 0.0021 \\
ASR Adaptation     & -- & -- & 0.9622 & 0.9957 & 0.9863 & 0.0172 & 0.1062 \\
\midrule
SITA               & 19 & \cmark & 0.9890 & 0.9994 & 0.9644 & 0.3500 & 0.6090 \\
SITA               & 19 & \xmark & 0.9896 & 0.9982 & 0.9650 & 0.3648 & 0.6491 \\
SITA               & 21 & \cmark & \textbf{0.9927} & \textbf{0.9994} & 0.9533 & 0.6469 & 1.0131 \\
SITA               & 21 & \xmark & 0.9915 & 0.9994 & 0.9530 & 0.6247 & 1.0042 \\
\bottomrule
\end{tabular}%
}
\vspace{-4pt}
\end{table}

\begin{table}[t]
\centering
\small
\setlength{\tabcolsep}{3.8pt}
\renewcommand{\arraystretch}{0.95}
\caption{Mandarin: ASR CER/WER. Lower is better.}
\label{tab:zh_asr_onecol}
\resizebox{\columnwidth}{!}{%
\begin{tabular}{lcc|cc}
\toprule
\multicolumn{3}{c|}{Setup} & \multicolumn{2}{c}{ASR (CTC)} \\
\cmidrule(lr){1-3}\cmidrule(lr){4-5}
Setting & Feat & Frz & CER/WER & CER/WER (+KD) \\
\midrule
Whisper (Large-v3)    & -- & -- & 0.5006/0.6787 & -- \\
OmnilingualASR 7B     & -- & -- & 1.0000/1.0000 & -- \\
ASR Adaptation        & -- & -- & 0.0107/0.0360 & -- \\
\midrule
SITA                  & 19 & \cmark & 0.0073/0.0274 & 0.0071/0.0280 \\
SITA                  & 19 & \xmark & 0.0086/0.0311 & 0.0078/0.0280 \\
SITA                  & 21 & \cmark & 0.0089/0.0323 & 0.0148/0.0555 \\
SITA                  & 21 & \xmark & 0.0082/0.0305 & 0.0144/0.0561 \\
\bottomrule
\end{tabular}%
}
\vspace{-4pt}
\end{table}

We transfer \textbf{SITA} to the Mandarin corpus using the same architecture, objectives, and hyperparameters as Hmong, changing only the training data and tone inventory. Table~\ref{tab:zh_retr_tone_onecol} summarizes retrieval and tone geometry results and Table~\ref{tab:zh_asr_onecol} summarizes ASR results. \textbf{SITA} achieves near-ceiling Male$\rightarrow$Female retrieval while maintaining strong tone structure.

\end{document}